\documentclass{article}

\usepackage{microtype}
\usepackage{graphicx}
\usepackage{booktabs} 
\usepackage{makecell}
\usepackage{graphicx}
\usepackage{caption}
\usepackage{courier}
\usepackage{multirow}
\usepackage{color}
\usepackage{subcaption}
\usepackage{color}
\usepackage{todonotes}
\usepackage{amssymb}
\usepackage{gensymb}
\usepackage{amsfonts}       
\usepackage{enumitem}
\usepackage{multirow}
\usepackage{amsmath,nccmath}

\newtheorem{theorem}{Theorem}
\usepackage{hyperref}

\expandafter\def\expandafter\normalsize\expandafter{%
     \normalsize
     \setlength\abovedisplayskip{2pt}
     \setlength\belowdisplayskip{2pt}
     \setlength\abovedisplayshortskip{2pt}
     \setlength\belowdisplayshortskip{2pt}
}
\usepackage[accepted]{icml2019}

\icmltitlerunning{Hierarchically Structured Meta-learning}

\begin{document}

\twocolumn[
\icmltitle{Hierarchically Structured Meta-learning}


\begin{icmlauthorlist}
\icmlauthor{Huaxiu Yao$^\dag$}{to}
\icmlauthor{Ying Wei}{goo}
\icmlauthor{Junzhou Huang}{goo}
\icmlauthor{Zhenhui Li}{to}
\end{icmlauthorlist}

\icmlaffiliation{to}{College of Information Science and Technology, Pennsylvania State University, PA, USA}
\icmlaffiliation{goo}{Tencent AI Lab, Shenzhen, China}

\icmlcorrespondingauthor{Ying Wei}{judyweiying@gmail.com}

\icmlkeywords{Meta-learning}

\vskip 0.3in
]

\printAffiliationsAndNotice{$^\dag$
Part of the work was done when the author interned in Tencent AI Lab.} 

\begin{abstract}
In order to learn quickly with few samples, meta-learning utilizes prior knowledge learned from previous tasks. However, a critical challenge in meta-learning is task uncertainty and heterogeneity, which can not be handled via globally sharing knowledge among tasks. In this paper, based on gradient-based meta-learning, we propose a hierarchically structured meta-learning (HSML) algorithm that explicitly tailors the transferable knowledge to different clusters of tasks.
Inspired by the way human beings organize knowledge, we resort to a hierarchical task clustering structure to cluster tasks.  
As a result, the proposed approach not only addresses the challenge via the knowledge customization to different clusters of tasks, but also preserves
knowledge generalization among a cluster of similar tasks.
To tackle the changing of task relationship, in addition, we extend the hierarchical structure to a continual learning environment. The experimental results show that our approach can achieve state-of-the-art performance in both toy-regression and few-shot image classification problems.
\end{abstract}
\section{Introduction}
Learning quickly with a few samples is one of the key characteristics of human intelligence, while it remains a daunting challenge for artificial intelligence.
Learning to learn (a.k.a., meta-learning)~\cite{braun2010structure}, as a common practice to address this challenge, leverages the transferable knowledge learned from previous tasks to improve the learning effectiveness in a new task.
There have been several lines of meta-learning algorithms, including recurrent network based methods~\cite{ravi2016optimization}, optimizer based methods~\cite{andrychowicz2016learning}, nearest neighbours based methods~\cite{snell2017prototypical,vinyals2016matching} and gradient descent based methods~\cite{finn2017model}, which instantiate the transferable knowledge as latent representations, an optimizer, a metric space, and parameter initialization, respectively.

Despite their early success in few-shot image classification~\cite{ravi2016optimization} and machine translation~\cite{gu2018meta}, most of the existing meta-learning algorithms assume the transferable knowledge to be globally shared across all tasks.
As a consequence, they suffer from handling a sequence of tasks originated from different distributions.
At the other end of the spectrum, recently, a few research works~\cite{finn2018probabilistic,yoon2018bayesian,lee2018gradient} try to fix the problem by tailoring the transferable knowledge to each task.
Yet the downside of such methods lies in the impaired knowledge generalization among closely correlated tasks (e.g., the tasks sampled from the same distribution). 

Hence we are motivated to pursue a meta-learning framework to effectively balance generalization and customization. 
The inspiration comes from a hypothesis which has been formulated and tested by psychological researchers~\cite{gershman2010context,gershman2014statistical}. 
The hypothesis suggests that the key to human beings' capability of solving a task with little training data is the way how human beings organize the learned knowledge from tasks.
As bits of tasks impinge on us, we human beings cluster the tasks into several states based on task similarity, so that the learning occurs within each cluster instead of across cluster boundaries.
Thus, when a new task arrives, it can either quickly take advantage of the knowledge learned within the cluster it belongs to or initiate a new cluster if it is wildly different from any existing clusters.

Inspired by this, we propose a novel meta-learning framework called \textbf{H}ierarchically \textbf{S}tructured \textbf{M}eta-\textbf{L}earning (HSML). 
The key idea of the HSML is to enhance the meta-learning effectiveness by
promoting knowledge customization to different clusters of tasks but simultaneously preserving knowledge generalization among a cluster of closely related tasks.
In this paper, without loss of generality, we ground HSML on a gradient based meta learning algorithm~\cite{finn2017model} with the transferable knowledge instantiated as parameter initializations. 
Specifically, first, the HSML resorts to a hierarchical clustering structure to perform soft clustering on tasks. 
The representation of each task is learned from either of the two proposed candidate aggregators, i.e., pooling autoencoder aggregator and recurrent autoencoder aggregator, and is passed to the hierarchical clustering structure to obtain the clustering result of this task.
The sequentially incoming tasks, in turn, update the clustering structure. Especially, if the existing structure does not fit the task, we dynamically expand the structure.
Secondly, a globally shared parameter initialization is tailored to each cluster via a parameter gate, to serve as the initializations for all tasks belonging to the cluster.

Again we would highlight the contribution of the proposed HSML: 1) it achieves a better balance between generalization and customization of the transferable knowledge, so that it empirically outperforms state-of-the-art meta-learning algorithms in both toy regression and few-shot image classification problems; 2) it is interpretable in terms of the task relationship; 3) it has been theoretically proved to be superior than existing gradient-based meta-learning algorithms.
\section{Related Work}
\begin{figure}[t]
	\centering
	\begin{subfigure}[c]{0.155\textwidth}
		\centering
		\includegraphics[height=21mm]{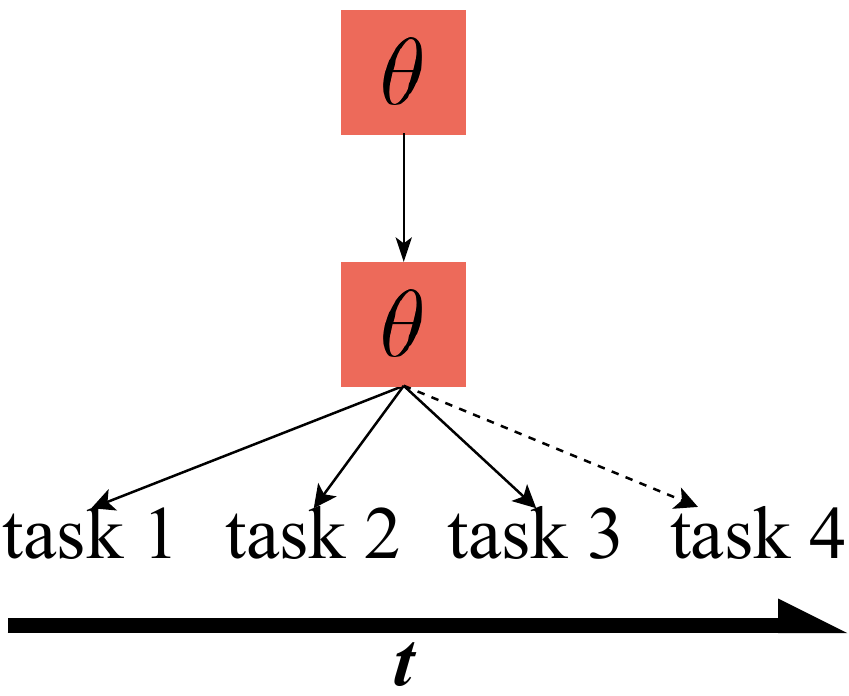}
		\caption{\label{fig:maml}: globally shared}
	\end{subfigure}
	\begin{subfigure}[c]{0.155\textwidth}
		\centering
		\includegraphics[height=21mm]{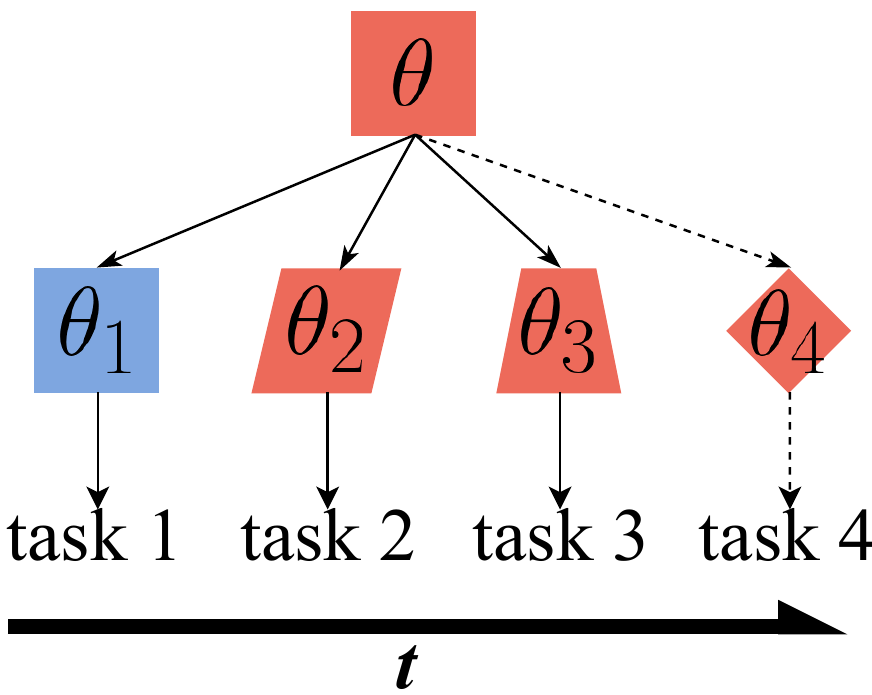}
		\caption{\label{fig:mumomaml}: task specific}
	\end{subfigure}
		\begin{subfigure}[c]{0.155\textwidth}
		\centering
		\includegraphics[height=21mm]{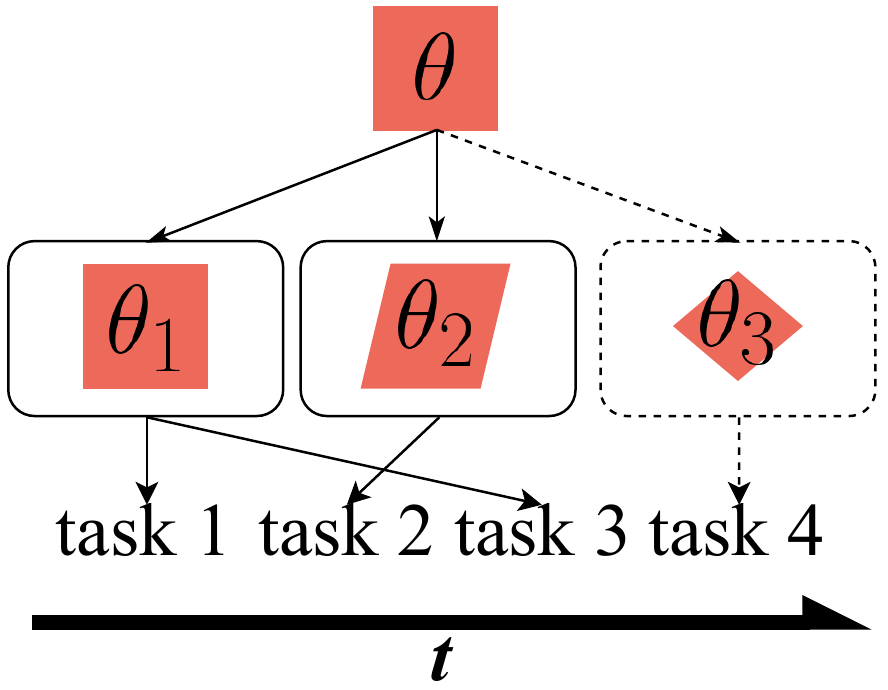}
		\caption{\label{fig:hsml}: HSML}
	\end{subfigure}
	\caption{Pictorial illustration of the difference between the proposed HSML and the other two representative lines of gradient based meta-learning algorithms.}
	
	\label{fig:illustration}
\end{figure}
Meta-learning, allowing machines to learn new skills or adapt to new environments rapidly with a few training examples, has demonstrated success in both supervised learning such as few-shot image classification and reinforcement learning settings.
There are four common approaches: 1) use a recurrent neural network equipped with either external or internal memory storing and querying meta-knowledge~\cite{munkhdalai2017meta,santoro2016meta,munkhdalai2018rapid,mishra2018simple}; 2) learn a meta-optimizer which can quickly optimize the model parameters~\cite{ravi2016optimization,andrychowicz2016learning,li2016learning}; 3) learn an effective distance metric between examples~\cite{snell2017prototypical,vinyals2016matching,yang2018learning};
4) learn an appropriate initialization from which the model parameters can be updated within a few gradient steps~\cite{finn2017model,finn2018probabilistic,lee2018gradient}.

HSML falls into the fourth aforementioned category named as gradient-based meta-learning.
Most of the gradient-based meta-learning algorithms~\cite{finn2017model,li2017meta,flennerhag2018transferring} assume a globally shared initialization across all tasks, as shown in Figure~\ref{fig:maml}. 
To accommodate dynamically changing tasks, as illustrated in Figure~\ref{fig:mumomaml}, recent  
studies tailor the global shared initialization to each task by taking advantage of 
probabilistic models~\cite{finn2018probabilistic,grant2018recasting,yoon2018bayesian} and 
incorporating task-specific information~\cite{lee2018gradient,vuorio2018toward}. 
However, our proposed HSML outlined in Figure~\ref{fig:hsml} customizes the global shared initialization to each cluster using a hierarchical clustering structure, which enjoys not only knowledge customization but also generalization (e.g., between task 1 and 3).
Better yet, HSML 
right fit to a continual learning scenario with evolving clustering structures.
\section{Preliminaries}
\label{sec:preliminaries}
\textbf{The Meta-Learning Problem} 
Suppose that a sequence of 
tasks \begin{small}$\{\mathcal{T}_1, ..., \mathcal{T}_{N_t}\}$\end{small} are sampled from an environment which is a probability distribution \begin{small}$\mathcal{E}$\end{small} on tasks~\cite{baxter1998theoretical}.
In each task \begin{small}$\mathcal{T}_i\!\sim\! \mathcal{E}$\end{small}, we have 
a few examples \begin{small}
$\{\mathbf{x}_{i,j},\mathbf{y}_{i,j}\}_{j=1}^{n^{tr}}$\end{small}
to constitute the training set \begin{small}$\mathcal{D}_{\mathcal{T}_i}^{tr}$\end{small}
and the rest 
as the test set \begin{small}$\mathcal{D}_{\mathcal{T}_i}^{te}$\end{small}.
Given a base learner \begin{small}$f$\end{small} with  \begin{small}$\theta$\end{small} as parameters, 
the optimal parameters \begin{small}
$\theta_{\mathcal{T}_i}$
\end{small} are learned 
to make accurate predictions, i.e., \begin{small}$f_{\theta_{\mathcal{T}_i}}(\mathbf{x}_{i,j})\!\rightarrow\!\mathbf{y}_{i,j}$\end{small}.
The effectiveness of 
such a base learner 
on \begin{small}$\mathcal{D}^{tr}_{\mathcal{T}_i}$\end{small} is evaluated by the loss function  \begin{small}$\mathcal{L}(f_{\theta_{\mathcal{T}_i}}, \mathcal{D}^{tr}_{\mathcal{T}_i})$\end{small}, 
which equals the mean square error \begin{small}$\sum_{(\mathbf{x}_{i,j}, \mathbf{y}_{i,j})\in \mathcal{D}^{tr}_{\mathcal{T}_i}}\Vert f_{\theta_{\mathcal{T}_i}}(\mathbf{x}_{i,j})\!-\!\mathbf{y}_{i,j}\Vert_2^2$\end{small} for regression problems 
or the cross entropy loss
\begin{small}$-\!\sum_{(\mathbf{x}_{i,j}, \mathbf{y}_{i,j})\in \mathcal{D}^{tr}_{\mathcal{T}_i}}\log p(\mathbf{y}_{i,j}|\mathbf{x}_{i,j}, f_{\theta_{\mathcal{T}_i}})$\end{small} for classification problems.   

The goal of meta-learning is to learn from previous tasks a well-generalized 
meta-learner \begin{small}$\mathcal{M}(\cdot)$\end{small} 
which 
can 
facilitate the training of the base learner in a future task with a few examples. 
In fulfillment of this, meta-learning involves two stages, i.e., meta-training and meta-testing.
During meta-training, the parameters of the base learner for all tasks, i.e., \begin{small}$\{\theta_{\mathcal{T}_i}\}_{i=1}^{N_t}$\end{small}, and the meta-learner \begin{small}$\mathcal{M}(\cdot)$\end{small} are optimized alternatingly.  
In virtue of \begin{small}$\mathcal{M}$\end{small},  the parameters \begin{small}$\{\theta_{\mathcal{T}_i}\}_{i=1}^{N_t}$\end{small} are learned to minimize the expected empirical loss over training sets of all \begin{small}$N_t$\end{small} historical tasks, i.e., \begin{small}$\min_{\{\theta_{\mathcal{T}_i}\}_{i=1}^{N_t}}\sum_{i=1}^{N_t} \mathcal{L}(\mathcal{M}(f_{\theta_{\mathcal{T}_i}}), \mathcal{D}^{tr}_{\mathcal{T}_i})$\end{small}.  
In turn, a well-generalized \begin{small}$\mathcal{M}$\end{small} can be obtained by minimizing the expected empirical loss over test sets, i.e., \begin{small}$\min_{\mathcal{M}}\sum_{i=1}^{N_t} \mathcal{L}(\mathcal{M}(f_{\theta_{\mathcal{T}_i}}), \mathcal{D}^{te}_{\mathcal{T}_i})$\end{small}.  
When it comes to the meta-testing phase, provided with 
a future task \begin{small}$\mathcal{T}_{t}$\end{small},  the learning effectiveness and efficiency 
are improved by applying the meta-learner 
\begin{small}$\mathcal{M}$\end{small} and solving  \begin{small}$\min_{\theta_{\mathcal{T}_{t}}}\mathcal{L}(\mathcal{M}(f_{\theta_{\mathcal{T}_{t}}}), \mathcal{D}^{tr}_{\mathcal{T}_{t}})$\end{small}.\\
\textbf{Gradient-based Meta-Learning} Here we give an overview of the representative algorithm,  model-agnostic meta-learning (MAML)~\cite{finn2017model}.
MAML instantiates the meta-learner \begin{small}$\mathcal{M}$\end{small}
as a well-generalized initialization for the parameters of a base learner from which a few gradient descent steps can be performed to reach the optimal \begin{small}$\theta_{\mathcal{T}_{i}}$\end{small} for the task \begin{small}$\mathcal{T}_i$\end{small}, which means
\begin{small}
$\mathcal{M}(f_{\theta_{\mathcal{T}_i}})=f_{\theta_0-\alpha\nabla_\theta \mathcal{L}(f_{\theta}, \mathcal{D}_{\mathcal{T}_i}^{tr})}$.
\end{small}
As a result, 
the optimization of \begin{small}
$\mathcal{M}$\end{small} during meta-training is formulated as (one gradient step as exemplary):
\begin{equation}
\small
    \min_{\theta_0} \sum_{i=1}^{N_t}\mathcal{L}(f_{\theta_0-\alpha\nabla_\theta \mathcal{L}(f_{\theta}, \mathcal{D}_{\mathcal{T}_i}^{tr})},\mathcal{D}_{\mathcal{T}_i}^{te}).
\label{eqn:maml}
\end{equation}
\section{Methodology}
\begin{figure*}[h]
	\centering
 	\includegraphics[width=0.8\textwidth]{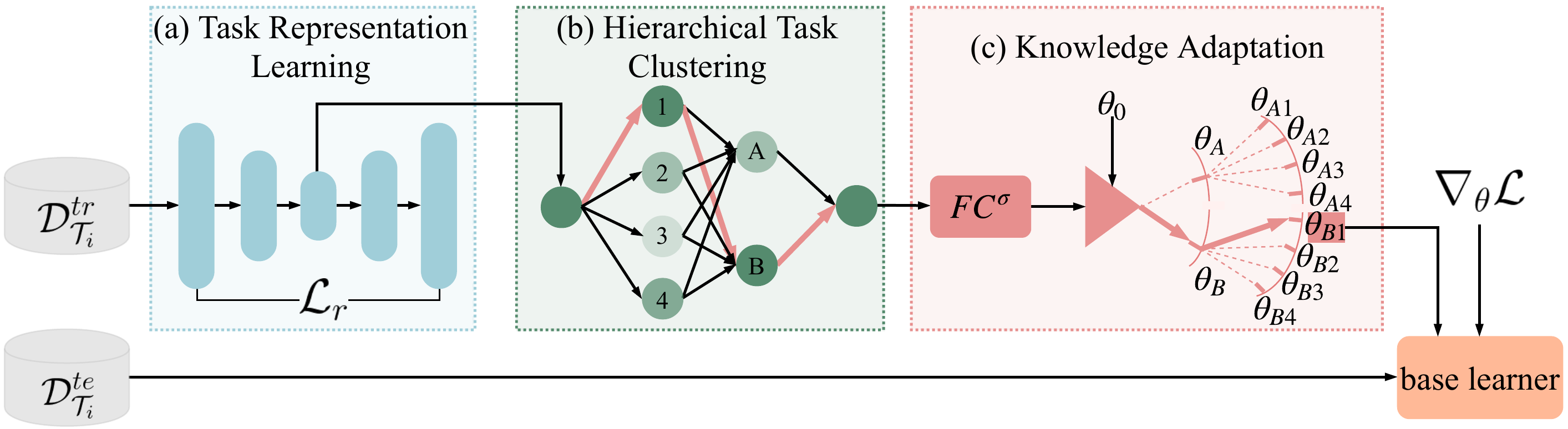}
	\caption{The framework of the proposed HSML involving three essential stages.
	(a) Task representation learning:
	we learn the representation for the task $\mathcal{T}_i$ using an autoencoder aggregator 
	(e.g., pooling aggregator, recurrent aggregator).
	(b) Hierarchical task clustering: provided with the task representation, we learn the soft clustering assignment with this differentiable hierarchical clustering structure. Darker nodes signify more likely assigned clusters (e.g., the cluster 1 in the first level and the cluster B in the second level).
	(c) Knowledge adaptation: we next use a parameter gate to adapt the transferable knowledge ($\theta_0$) to a cluster-specific initialization ($\theta_{B1}$) from which only a few gradient descent steps are required to achieve the optimal parameters $\theta_{\mathcal{T}_i}$.
	}
	\label{fig:framework}
\end{figure*}
In this section, we 
detail the proposed HSML 
algorithm whose framework is presented in Figure~\ref{fig:framework}. 
The HSML aims to adapt the transferable knowledge learned from previous tasks, namely the initialization for parameters of the base learner in gradient based meta-learning ($\theta_0$ here), to the task $\mathcal{T}_i$ in a cluster-specific manner, so that the optimal parameters $\theta_{\mathcal{T}_i}$ can be achieved in as few gradient descent steps as possible. 
As shown in the part (c) of Figure~\ref{fig:framework}, the possibilities to adaptation are completely dictated by the hierarchical task clustering structure in part (b), and the eventual path for adaptation follows the clustering result on the task $\mathcal{T}_i$ (i.e., $\theta_{B1}$).
By this means, the HSML balances between customization and generalization: the transferable knowledge is adapted to different clusters of tasks, while it is still shared among closely related tasks pertaining to the same cluster. 
To perform hierarchical clustering on tasks, we learn the representation of a task using the proposed task embedding network, i.e., the part (a). 
Next we will introduce the three stages, i.e., \emph{task representation learning}, \emph{hierarchical task clustering}, and \emph{knowledge adaptation}, respectively.

\subsection{Task Representation Learning}
Learning the representation of a task \begin{small}$\mathcal{T}_i$\end{small} with
the whole training set \begin{small}$\mathcal{D}^{tr}_{\mathcal{T}_i}$\end{small} as input is much more challenging than 
the common representation learning over examples,
which bears a striking similarity to the connection between sentence embeddings and word embeddings in natural language processing. 
Inspired by common practices in learning sentence embeddings~\cite{conneau2017supervised}, we tackle the challenge by
aggregating representations of all examples \begin{small}$\{\mathbf{x}_{i,j},\mathbf{y}_{i,j}\}_{j=1}^{n^{tr}}\!\in\!\mathcal{D}^{tr}_{\mathcal{T}_i}$\end{small}.
The desiderata of 
an ideal aggregator include 1) high 
representational capacity, and 2) permutational invariance 
to its inputs. 
In light of these, we propose two candidate aggregators, i.e., pooling autoencoder aggregator (PAA) and recurrent autoencoder aggregator (RAA).  
\\
\textbf{Pooling Autoencoder Aggregator} 
To meet the first desideratum, foremost, we resort to an autoencoder that learns highly effective representation for each example. 
The recontruction loss for training the autoencoder is as follows,
\begin{equation}
\small
\label{eq:reconstructionloss}
    \mathcal{L}_{r}(\mathcal{D}_{\mathcal{T}_i}^{tr})=\sum_{j=1}^{n^{tr}}\Vert \mathrm{FC}_{dec}(\mathbf{g}_{i,j})-\mathcal{F}(\mathbf{x}^{tr}_{i,j}, \mathbf{y}^{tr}_{i,j})\Vert_2^2,
\end{equation}
where \begin{small} $\mathbf{g}_{i,j} \!=\! \mathrm{FC}_{enc}(\mathcal{F}(\mathbf{x}^{tr}_{i,j}, \mathbf{y}^{tr}_{i,j}))$ \end{small} is the representation for the $j$-th example. 
In order to characterize the joint distribution instead of the marginal distribution only, we use \begin{small}$\mathcal{F}(\cdot,\cdot)$\end{small} to preliminarily embed both features and predictions of an example. 
The definition of \begin{small}$\mathcal{F}$\end{small} varies from dataset to dataset, which we will
detail in supplementary material C.
\begin{small}$\mathrm{FC}_{enc}$\end{small} and \begin{small}$\mathrm{FC}_{dec}$\end{small} stand for the encoder composed of a stack of fully connected layers and the decoder consisting of two fully connected layers with ReLU activation, respectively. 
Consequently, the aggregation satisfying the permutational invariance follows,
\begin{equation}
\label{eq:pool}
\small
    \mathbf{g}_i=\mathrm{Pool}_{j=1}^{n^{tr}}(\mathbf{g}_{i,j}),
\end{equation}
where \begin{small}$\mathbf{g}_i\!\in\! \mathbb{R}^d$\end{small} is the desired representation of task \begin{small}$\mathcal{T}_i$\end{small}. \begin{small}$\mathrm{Pool}$\end{small} denotes a max or mean pooling operator 
over examples. 
\\
\textbf{Recurrent Autoencoder Aggregator}
Motivated by recent success of the recurrent embedding aggregation in order-invariant problems such as graph embedding ~\cite{hamilton2017inductive}, we also consider a recurrent autoencoder aggregator which demonstrates more remarkable expressivity especially for a task with few examples.
Different from the pooling autoencoder, examples are sequentially fed into the recurrent autoencoder, i.e.,
\begin{equation}
\small
    \mathcal{F}(\mathbf{x}^{tr}_{i,1}, \mathbf{y}^{tr}_{i,1})
    \!\rightarrow\!
    \mathbf{g}_{i,1}
    \!\rightarrow\!\cdots\!\rightarrow\! \mathbf{g}_{i,n^{tr}}\!\rightarrow\!\mathbf{d}_{i,n^{tr}}
    \!\rightarrow\!
    \cdots
    \!\rightarrow\!
    \mathbf{d}_{i,1},
\end{equation}
where \begin{small}
$\forall j, \mathbf{g}_{i,j}\!=\!\mathrm{RNN}_{enc}(\mathcal{F}(\mathbf{x}^{tr}_{i,j}, \mathbf{y}^{tr}_{i,j}), \mathbf{g}_{i,j-1})$
\end{small}
and \begin{small}
$ \mathbf{d}_{i,j}\!=\!\mathrm{RNN}_{dec}(\mathbf{d}_{i,j+1})$
\end{small}
represent the learned representation  and the reconstruction of the $j$-th example, respectively.
Here
\begin{small}
$\mathrm{RNN}_{enc}$
\end{small}
and \begin{small}
$\mathrm{RNN}_{dec}$
\end{small}
stand for a recurrent encoder (LSTM or GRU) and a recurrent decoder, respectively.
The reconstruction loss is similar to Eqn.~\eqref{eq:reconstructionloss}, except that \begin{small}$\mathrm{FC}_{dec}(\mathbf{g}_{i,j})$\end{small} is replaced with \begin{small}$\mathbf{d}_{i,j}$\end{small}.
Thereupon, the task representation is aggregated over representations of all examples, i.e.,
\begin{equation}
\label{eq:rnn}
\small
    \mathbf{g}_i=\frac{1}{n^{tr}}\sum_{j}^{n^{tr}}(\mathbf{g}_{i,j}).
\end{equation}
Regrettably, the sequential feeding of examples makes the final task representation to be permutation sensitive, which violates the second prerequisite of an ideal aggregator. 
We address the problem by 
applying the recurrent aggregator to random permutations of examples~\cite{hamilton2017inductive}. 
\subsection{Hierarchical Task Clustering}
\label{sec:relationstructure}
Given the representation of a task, we propose a hierarchical task clustering structure to locate the cluster the task belongs to.
Before proceeding to detail the structure, we first explicate why the hierarchical clustering 
is preferred over flat clustering: a single level of task groups is likely insufficient to model complex task relationship in real-world applications; 
for example, to identify the cross-talks between gene expressions of multiple species, the study~\cite{kim2010tree} suggests multi-level clustering of such gene interaction.

The hierarchical clustering, following the tradition of clustering, proceeds by alternating between two steps, i.e., 
assignment step and update step, in a layer-wise manner. \\
\textbf{Assignment step}: 
Each task 
receives a cluster assignment score on each 
hierarchical level, and the assignment that it receives in a particular level is a function of its representation in the previous level. Thus, we assign a task represented in the $k^l$-th cluster of the $l$-th level, i.e., 
\begin{small}
$
\mathbf{h}^{k^l}_i\!\in\mathbb{R}^d\!
$
\end{small},
to the $k^{l+1}$-th cluster in the $(l\!+\!1)$-th level.
Note that we conduct soft assignment for the following two reasons: (1) task groups have been demonstrated to 
overlap, since there is always a 
continuum in the sharing between tasks~\cite{kumar2012learning}; (2) the soft instead of hard assignment guarantees the differentiability, so that the full HSML
framework can still be trained in an end-to-end fashion. In particular, for each task \begin{small}$\mathcal{T}_i$\end{small}, the soft-assignment \begin{small}$p_{i}^{k^l\rightarrow k^{l+1}}$\end{small} 
is computed by
applying softmax over Euclidean distances between \begin{small}
$
\mathbf{h}^{k^l}_i
$
\end{small} and the learnable cluster centers \begin{small}$\{\mathbf{c}_{k^{l+1}}\}_{k^{l+1}=1}^{K^{l+1}}$\end{small}, i.e., 
\begin{equation}
\label{eq:cluster_assign}
\small
p_{i}^{k^l\rightarrow k^{l+1}} = \frac{\exp{(-\Vert (\mathbf{h}^{k^l}_i - \mathbf{c}_{k^{l+1}})/\sigma^l\Vert_2^2/2)}}{\sum_{k^{l+1}=1}^{K^{l+1}} \exp{(-\Vert (\mathbf{h}^{k^l}_i - \mathbf{c}_{k^{l+1}})/\sigma^l\Vert_2^2}/2)},
\end{equation}
where $\sigma^l$ is a scaling factor in the $l$-th level and $K^{l+1}$  denotes the number of clusters in the $(l\!+\!1)$-th level. 
\\
\textbf{Update step}: 
As a result of assignment, the representation of a task in the $k^{l+1}$-th cluster of the $(l\!+\!1)$-th level, i.e., 
\begin{small}
$
\mathbf{h}^{k^{l+1}}_i
$,
\end{small}
can be updated with the following weighted average,
\begin{equation}
\small
\label{eq:cluster_repr}
    \mathbf{h}_i^{k^{l+1}} = \sum_{k^{l}=1}^{K^l} p^{k^{l}\rightarrow k^{l+1}}_i\tanh{(\mathbf{W}^{k^{l+1}}\mathbf{h}_i^{k^{l}}+\mathbf{b}^{k^{l+1}})}, 
\end{equation}
where \begin{small}$\mathbf{W}^{k^{l+1}}\!\in\mathbb{R}^{d\times d}\!$\end{small} and \begin{small}$\mathbf{b}^{k^{l+1}}\!\in\!\mathbb{R}^d$\end{small} are learned to transform from representations of the $l$-th to those of the $(l\!+\!1)$-th level.

The full pipeline of clustering starts from \begin{small}$l\!=\!0$\end{small}, where the initialization for 
\begin{small}
$
\mathbf{h}^{k^0}_i
$
\end{small}
equals the task representation \begin{small}$\mathbf{g}_i$\end{small} and \begin{small}
$K^0\! =\! 1$,
\end{small}
and ends at \begin{small}
$K^L\! =\! 1$.
\end{small}
We would especially discuss the cluster centers. 
The meta-learning scenario where training tasks come sequentially poses a unique challenge 
which requires the hierarchical clustering structure to be accordingly online. 
Therefore, the cluster centers are parameterized and learned as the learning proceeds. Each center is randomly initialized.
\subsection{Knowledge Adaptation}
The final representation 
\begin{small}
$\mathbf{h}_i^L$
\end{small}, which encrypts the hierarchical clustering result, 
is believed to be cluster specific.
Previous works~\cite{xu2015show,lee2018gradient} suggest that similar tasks activate similar meta-parameters 
(e.g., initialization) while different tasks trigger disparate ones. Inspired by this finding, we 
design a cluster-specific parameter gate, 
\begin{equation}
\small
\label{eq:gate}
    \mathbf{o}_i=\mathrm{FC}^{\sigma}_{\mathbf{W}_g}(\mathbf{g}_i\oplus \mathbf{h}_i^L),
\end{equation}
where the fully connected layer \begin{small}$\mathrm{FC}^{\sigma}_{\mathbf{W}_g}$\end{small} 
is parameterized by \begin{small}$\mathbf{W}_g$\end{small} and activated by a sigmoid function $\sigma$.
It is worth mentioning here that concatenating the task representation \begin{small}
$\mathbf{g}_i$
\end{small} together with \begin{small}
$\mathbf{h}_i^L$
\end{small}
not only preserves but also reinforces the cluster-specific property of the parameter gate. Most importantly, the globally transferable knowledge, i.e., the initial parameters $\theta_0$, is adapted to the cluster-specific initial parameters $\theta_{0i}$ via the parameter gate, i.e., 
\begin{small}$\theta_{0i}\!=\!\theta_0\circ \mathbf{o}_i$\end{small}.

Recalling the objectives for a meta-learning algorithm in Section~\ref{sec:preliminaries}, we reach the optimization problem for 
HSML: 
\begin{equation}
\small
\label{eq:lossall}
    \min_{\Theta} \sum_{i=1}^{ N_t}\mathcal{L}(f_{\theta_{0i}-\alpha\nabla_{\theta} \mathcal{L}(\theta, \mathcal{D}_{\mathcal{T}_i}^{tr})},\mathcal{D}_{\mathcal{T}_i}^{te})+\xi\mathcal{L}_{r}(\mathcal{D}_{\mathcal{T}_i}^{tr}),
\end{equation}
where $\mathcal{L}$ defined in Section~\ref{sec:preliminaries} measures the empirical risk 
over $\mathcal{D}^{te}_{\mathcal{T}_i}$ and $\mathcal{L}_{r}$ measures the reconstruction error as defined in Eqn.~\eqref{eq:reconstructionloss}.
$\xi$ is used to balance the importance of these two items. 
$\Theta$ represents 
all learnable parameters including the global transferable initialization $\theta_0$, the parameters for clustering, and those for knowledge adaptation (i.e., \begin{small}$\mathbf{W}_g$\end{small}).
\\
\textbf{Continual Adaptation}
We especially pay attention to the case where a new task does not fit any of the learned task clusters, which implies that 
additional clusters 
should be introduced to the 
hierarchical clustering 
structure. Incrementally adding model capacity~\cite{yoon2017lifelong,daniely2015strongly}, 
has been the common practice to handle distribution drift 
without initially introducing excessive
parameters. 
The key lies in the criterion when to expand the clustering structure. 
Since the loss values of \begin{small}
$\mathcal{L}(f_{\theta_{\mathcal{T}_i}}, \mathcal{D}^{te}_{\mathcal{T}_i})$
\end{small}
fluctuate 
across different tasks 
during the online meta-training process, setting the loss value as threshold would obviously be futile.
Instead, for every $Q$ training tasks, we compute the average loss value \begin{small}$\bar{\mathcal{L}}$\end{small}. 
If the new average value \begin{small}$\bar{\mathcal{L}}_{new}$\end{small} is more than \begin{small}$\mu$\end{small} times the previous value \begin{small}$\bar{\mathcal{L}}_{old}$\end{small} (i.e., \begin{small}$\bar{\mathcal{L}}_{new}>\mu\bar{\mathcal{L}}_{old}$\end{small}), the number of clusters will be increased, and the parameters for new clusters are randomly initialized. The whole algorithm of our proposed model is detailed in Alg.~\ref{alg:hsml}.
\begin{algorithm}[tb]
   \caption{Meta-training of HSML}
   \label{alg:hsml}
\begin{algorithmic}[1]
\REQUIRE \begin{small}$\mathcal{E}$\end{small}: distribution over tasks; \begin{small}$\{K^1,\cdots, K^L\}$\end{small}: 
\# of clusters in each layer; \begin{small}$\alpha$\end{small}, \begin{small}$\beta$\end{small}: stepsizes; \begin{small}$\mu$\end{small}: threshold
    \STATE Randomly initialize \begin{small}$\Theta$ 
    \end{small}
    \WHILE{not done}
    \IF{\begin{small}$\bar{\mathcal{L}}_{new}>\mu\bar{\mathcal{L}}_{old}$\end{small}}
    \STATE Increase the number of clusters 
    \ENDIF
    \STATE Sample a batch of tasks \begin{small}$\mathcal{T}_i\sim \mathcal{E}$\end{small}
    \FORALL{\begin{small}$\mathcal{T}_i$\end{small}}
    \STATE Sample \begin{small}$\mathcal{D}_{\mathcal{T}_i}^{tr}$\end{small}, \begin{small}$\mathcal{D}_{\mathcal{T}_i}^{te}$\end{small} from \begin{small}$\mathcal{T}_i$\end{small}
    \STATE Compute \begin{small}$\mathbf{g}_i$\end{small} in Eqn.~\eqref{eq:pool} or Eqn.~\eqref{eq:rnn}, $\mathbf{h}_i^L$ in Eqn.~\eqref{eq:cluster_repr}, and reconstruction error \begin{small}$\mathcal{L}_r(\mathcal{D}_{\mathcal{T}_i}^{tr})$\end{small}
    \STATE Compute \begin{small}$\mathbf{o}_i$\end{small} in Eqn.~\eqref{eq:gate} and evaluate \begin{small}$\nabla_\theta \mathcal{L}(\theta,\mathcal{D}^{tr}_{\mathcal{T}_i})$\end{small}
    \STATE Update parameters with gradient descent (taking one step as an example): \begin{small}$\theta_{\mathcal{T}_i}\!=\!\theta_{0i}\!-\!\alpha\nabla_{\theta} \mathcal{L}(\theta, \mathcal{D}_{\mathcal{T}_i}^{tr})$\end{small}
    \ENDFOR
    \STATE Update \begin{small}$\Theta\leftarrow \Theta-\beta\nabla_{\Theta}\sum_{i=1}^{N_t}\mathcal{L}(f_{\theta_{\mathcal{T}_i}},\mathcal{D}_{\mathcal{T}_i}^{te})+\xi\mathcal{L}_{r}(\mathcal{D}_{\mathcal{T}_i}^{tr})$\end{small} 
    \STATE Compute \begin{small}$\bar{\mathcal{L}}_{new}$\end{small} and save \begin{small}$\bar{\mathcal{L}}_{old}$\end{small} for every \begin{small}$Q$\end{small} rounds
    \ENDWHILE
\end{algorithmic}
\end{algorithm}

\section{Analysis}
The 
core 
of HSML is to adapt a 
globally shared initialization of stochastic gradient descent (SGD) to be cluster specific via the proposed hierarchical clustering structure. 
Hence, in this section, we theoretically analyze the advantage of such adaptation in terms of the generalization bound. 

For a task \begin{small}$\mathcal{T}_{i}\!\sim\!\mathcal{E}$\end{small}, we assume 
both training and testing examples are i.i.d. drawn from a distribution \begin{small}$\mathcal{S}_{i}$\end{small} 
, i.e., \begin{small}$\mathcal{D}_{\mathcal{T}_i}^{tr}
\!\sim\! \mathcal{S}_{i}$\end{small} and \begin{small}$\mathcal{D}_{\mathcal{T}_i}^{te}
\!\sim\! \mathcal{S}_{i}$\end{small}. 
According to 
Theorem 2 in~\cite{kuzborskij2017data}, a base learner 
\begin{small}$f_{\theta_{\mathcal{T}_i}}$\end{small} is \begin{small}$\epsilon(\mathcal{S}_i,\theta_0)$\end{small}-on-average stable if its generalization is bounded by \begin{small}$\epsilon(\mathcal{S}_i,\theta_0)$\end{small}, i.e., \begin{small}$\mathbb{E}_{\mathcal{S}_i}\mathbb{E}_{f_{\theta_{\mathcal{T}_i}}}[R(f_{\theta_{\mathcal{T}_i}}(\mathcal{D}^{tr}_{\mathcal{T}_i}))\!-\!\hat{R}_{\mathcal{D}^{tr}_{\mathcal{T}_i}}(f_{\theta_{\mathcal{T}_i}}(\mathcal{D}^{tr}_{\mathcal{T}_i}))]\!\leq\! \epsilon(\mathcal{S}_i,\theta_0)$\end{small}. \begin{small}$\theta_0$\end{small} is the initialization of SGD to reach \begin{small}$\theta_{\mathcal{T}_i}$\end{small}, and \begin{small}$R(\cdot)$\end{small} and \begin{small}$\hat{R}_{\mathcal{D}_{\mathcal{T}_i}^{tr}}(\cdot)$\end{small} denote the expected and empirical risk on \begin{small}
$\mathcal{D}_{\mathcal{T}_i}^{tr}$
\end{small}, respectively. 

Transferring the globally shared initialization \begin{small}$\theta_0$\end{small} (i.e., MAML) to the target task
\begin{small}$\mathcal{T}_{t}$\end{small} 
is equivalent to transferring a hypothesis \begin{small}$f_{\theta_0}$\end{small} learned from meta-training tasks
like~\cite{kuzborskij2017fast}. For HSML, the initialization can be represented as 
\begin{small}$\theta_{0t}=\sum_{k=1}^{K}\hat{\mathbf{B}}_k\theta_0$\end{small}, which we demonstrate in the supplementary material A. In the following two theorems, provided with an initialization \begin{small}
$\theta_{0t}$
\end{small}, we derive according to~\cite{kuzborskij2017data}  the generalization bounds of the base learner \begin{small}
$f_{\theta_{\mathcal{T}_t}}$
\end{small}
when the loss \begin{small}$\mathcal{L}$\end{small} is convex and non-convex, respectively. 
\begin{theorem}
\emph{Assume that \begin{small}$\mathcal{L}$\end{small} is convex and \begin{small}
$f_{\theta_{\mathcal{T}_t}}$
\end{small} optimized using SGD 
is \begin{small}$\epsilon\big(\mathcal{S}_{t}, \theta_{0t}\big)$\end{small}-on-average stable. Then \begin{small}$\epsilon\big(\mathcal{S}_{t},\theta_{0t}\big)$\end{small} is bounded by},
\begin{equation}
\small
\mathcal{O}\bigg(\sqrt{\hat{R}_{\mathcal{D}^{tr}_{\mathcal{T}_t}}(\theta_{0t})+\sqrt{\frac{1}{n^{tr}}}}\bigg).
\end{equation}
\end{theorem}
\begin{theorem}
\emph{Assume that \begin{small}$\mathcal{L}\in [0,1]$\end{small} is \begin{small}$\eta$\end{small}-smooth and has a \begin{small}$\rho$\end{small}-Lipschitz Hessian. The step size at the \begin{small}$u$\end{small}-step
\begin{small}$\alpha_u=c/u$\end{small} satisfying \begin{small}$c\leq\min\{\frac{1}{\eta},\frac{1}{4(2\eta\ln{U})^2}\}$\end{small} with total steps \begin{small}$U=n^{tr}$\end{small} and \begin{small}$\hat{\gamma}^{\pm}=\frac{1}{{n^{tr}}}\sum_{j=1}^{n^{tr}}\Vert\nabla^2\mathcal{L}(\theta_{0t}, (\mathbf{x}_{t,j},\mathbf{y}_{t,j}))\Vert_2+\sqrt{\hat{R}_{\mathcal{D}_{\mathcal{T}_t}^{tr}}(\theta_{0t})}\pm\frac{1}{\sqrt[4]{{n^{tr}}}}$\end{small} and then \begin{small}$\epsilon(\mathcal{S}_{t},\theta_{0t})$\end{small} is bounded by},
\begin{equation}
\small
\mathcal{O}\bigg(\Big(1+\frac{1}{c\hat{\gamma}^{-}}\Big)\hat{R}_{\mathcal{D}^{tr}_{\mathcal{T}_t}}(\theta_{0t})^{\frac{c\hat{\gamma}^{+}}{1+c\hat{\gamma}^{+}}}\frac{1}{{(n^{tr})}^{\frac{1}{1+c\hat{\gamma}^{+}}}}\bigg).
\end{equation}
\end{theorem}
Though some standard base learners (e.g., 4 convolutional layers in few-shot image classification~\cite{finn2017model}) with ReLU do not meet the property of Lipschitz Hessian, following~\cite{nguyen2018optimization}, a softplus function  \begin{small}$f(x)=\frac{1}{\kappa}\log(1+\exp(\kappa x))$\end{small} can arbitrarily well approximate ReLU by adjusting \begin{small}$\kappa$\end{small} and thus Theorem 2 holds. In both cases, 
MAML can be regarded as 
the special case of HSML, i.e., \begin{small}$\forall k, \hat{\mathbf{B}}_k=\mathbf{I}$\end{small}, where \begin{small}$\mathbf{I}$\end{small} is an identity matrix. 
Remarkably, by proving \begin{small}$\exists \{\hat{\mathbf{B}}_k\}_{k=1}^K, s.t.,\;\hat{R}_{\mathcal{D}^{tr}_{\mathcal{T}_t}}(\theta_{0t})\leq \hat{R}_{\mathcal{D}^{tr}_{\mathcal{T}_t}}(\theta_{0})$\end{small},
we conclude that HSML achieves a tighter generalization bound than MAML and thereby is much more favored.
Consider the optimization process starting from $\theta_{0}$, through the negative gradient direction, \begin{small}$\hat{\theta}_{0}\!=\!(\mathbf{I}-\alpha\nabla \mathcal{L}(\theta_{0})(\theta_{0}\mathbf{I})^{-1})\theta_{0}$\end{small} and \begin{small}$\hat{R}_{\mathcal{D}^{tr}_{\mathcal{T}_t}}(\mathcal{\hat{\theta}}_{0})\leq\hat{R}_{\mathcal{D}^{tr}_{\mathcal{T}_t}}(\mathcal{\theta}_{0})$\end{small}. Thus, we can find a \begin{small}$\sum_{k=1}^{K}\hat{\mathbf{B}}_k=\mathbf{I}-\alpha\nabla \mathcal{L}(\theta_{0})(\theta_{0}\mathbf{I})^{-1}$\end{small}. We provide more details about analysis in supplementary material A.
\section{Experiments}
\label{sec:exp}
In this section, we evaluate the effectiveness of HSML. The goal of our experimental evaluation is to answer the following questions: (1) Can our approach outperform other meta-learning algorithms in toy regression and few-shot image classification tasks? (2) Can our approach discover reasonable task clusters? (3) Can our approach update the clustering structure in the continual learning manner and achieve better performance?

We study these questions on toy regression and few-shot image classification problems. For gradient-based meta-learning algorithms, we select the following as baselines: (1) globally shared models including MAML~\cite{finn2017model} and Meta-SGD~\cite{li2017meta}; (2) task specific models including MT-Net~\cite{lee2018gradient}, BMAML~\cite{yoon2018bayesian} and MUMOMAML~\cite{vuorio2018toward}. 
The empirical results indicate that recurrent autoencoder aggregator (RAA) is 
on average better than PAA 
for task representation, 
so that RAA is used as the default aggregator. We also provide a comparison of RAA and PAA on few-shot classification problem in supplementary material G. All the baselines use the same neural network structure (base learner). For hierarchical task clustering, like~\cite{ying2018hierarchical}, the number of clusters in a high layer is half of that in its consecutive lower layer. We specify the hyperparameters 
for meta-training in supplementary material C. 
\subsection{Toy Regression}
\label{sec:toy_res}
\textbf{Dataset and Experimental Settings}
In the toy regression problem, different tasks are sampled from different family of functions. In this paper, the underlying family functions are (1) \emph{Sinusoids}: $y(x)\!=\!A\mathrm{sin}(wx)\!+\!b$, $A\!\sim\! U[0.1, 5.0]$, $w\!\sim\! U[0.8, 1.2]$ and $b\!\sim\! U[0, 2\pi]$; (2) \emph{Line}: $y(x)\!=\!A_lx+b_l$, $A_l\!\sim\! U[-3.0, 3.0]$ and $b_l\!\sim\! U[-3.0, 3.0]$; (3) \emph{Cubic}: $y(x)\!=\!A_c x^3\!+\!b_c x^2\! +\! c_c x \!+ \!d_c$, $A_c\!\sim\! U[-0.1,0.1]$, $b_c\!\sim\! U[-0.2, 0.2]$, $c_c\!\sim\! U[-2.0, 2.0]$ and $d_c\!\sim\! U[-3.0, 3.0]$; (4) \emph{Quadratic}: $y(x)\!=\!A_qx^2\!+\!b_q x\!+\! c_q$, $A_q\!\sim\! U[-0.2,0.2]$, $b_q\!\sim\! U[-2.0, 2.0]$ and $c_q\!\sim\! U[-3.0, 3.0]$. $U[\cdot,\cdot]$ represents a uniform distribution. Each individual is randomly sampled from one of the four underlying functions. The input $x\!\sim\! U[-5.0, 5.0]$ for both training and testing tasks. We train all models for 5-shot and 10-shot regression. Mean square error (MSE) is used as evaluation metric. In hierarchical clustering, we set the number of layers to be three with 4, 2, 1 clusters in each layer, respectively.
\\
\textbf{Results of Regression Performance}
The results of 5-shot and 10-shot regression are shown in Table~\ref{tab:1dregression_group}. 
HSML improves the performance of global models (e.g., MAML) and task specific models (e.g., MUMOMAML), indicating the effectiveness of task clustering. 
\begin{table}[h]
\small
\caption{Performance of MSE $\pm$ 95\% confidence intervals on toy regression tasks, averaged over 4,000 tasks. Both 5-shot and 10-shot results are reported.}
\label{tab:1dregression_group}
\begin{center}
\begin{tabular}{l|c|c}
\hline
Model & 5-shot &  10-shot \\\hline
MAML    & $2.205\pm0.121$ & $0.761\pm0.068$\\
Meta-SGD & $2.053\pm 0.117$ & $0.836\pm0.065$ \\
MT-Net & $2.435\pm 0.130$ & $0.967\pm0.056$\\
BMAML & $2.016\pm 0.109$ & $0.698\pm 0.054$\\
MUMOMAML & $1.096\pm 0.085$ & $0.256\pm 0.028$\\\midrule
\textbf{HSML (ours)}    & $\mathbf{0.856\pm 0.073}$ & $\mathbf{0.161\pm 0.021}$ \\\hline
\end{tabular}
\end{center}
\end{table}
\\
\textbf{Task Clustering Analysis in Toy Regression}
\begin{figure}[h]
	\centering
 	\includegraphics[width=0.46\textwidth]{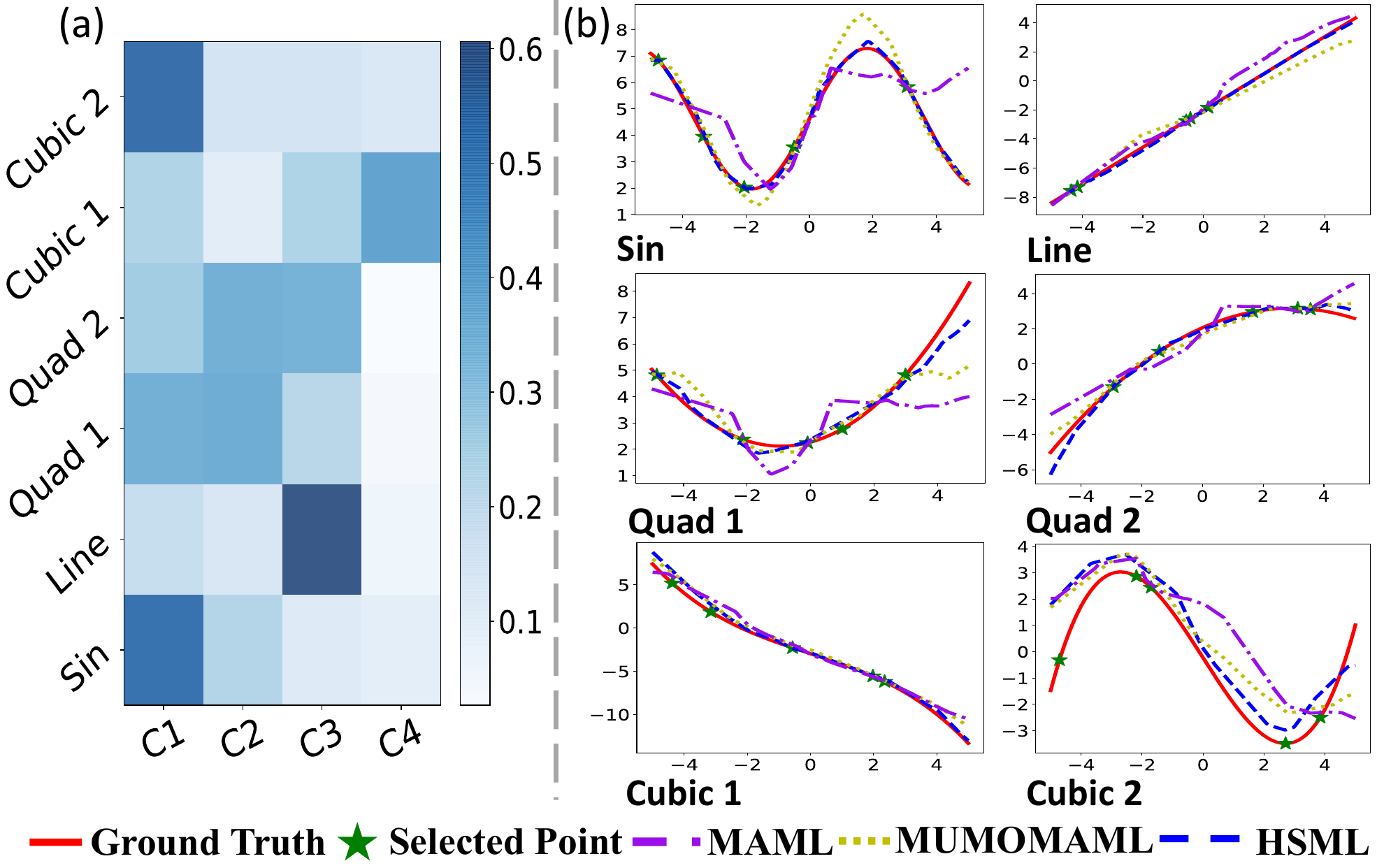}
	\caption{(a) The visualization of soft-assignment in Eqn.~\eqref{eq:cluster_assign} of six selected tasks. Darker color represents higher probability. (b) The corresponding fitting curves. 
	The ground truth underlying a function is shown in red line with data samples marked as green stars. C1-4 mean cluster 1-4, respectively.} 
	\label{fig:vis_sync}
\end{figure}
In order to show the power of HSML for detecting task clusters, we randomly select six tasks (more results are shown in supplementary material I) of 5-shot regression scenario and show soft-assignment values in Figure~\ref{fig:vis_sync}(a), i.e., the value of $\{p_i^{k^0\rightarrow k^1}|\forall k^1\}$ in Eqn.~\eqref{eq:cluster_assign}. Darker color stands for higher probability. 
The qualitative results of each task are shown in Figure~\ref{fig:vis_sync}(b). The ground truth underlying functions and the data samples $\mathcal{D}^{tr}_{\mathcal{T}_i}$ are shown as red lines and green stars, respectively. Qualitative results of MAML, MUMOMAML (best baseline), HSML are shown in different colors. 

\begin{figure}[h]
	\centering
	\begin{subfigure}[b]{0.9\linewidth}
	\centering
		\includegraphics[width=0.9\textwidth]{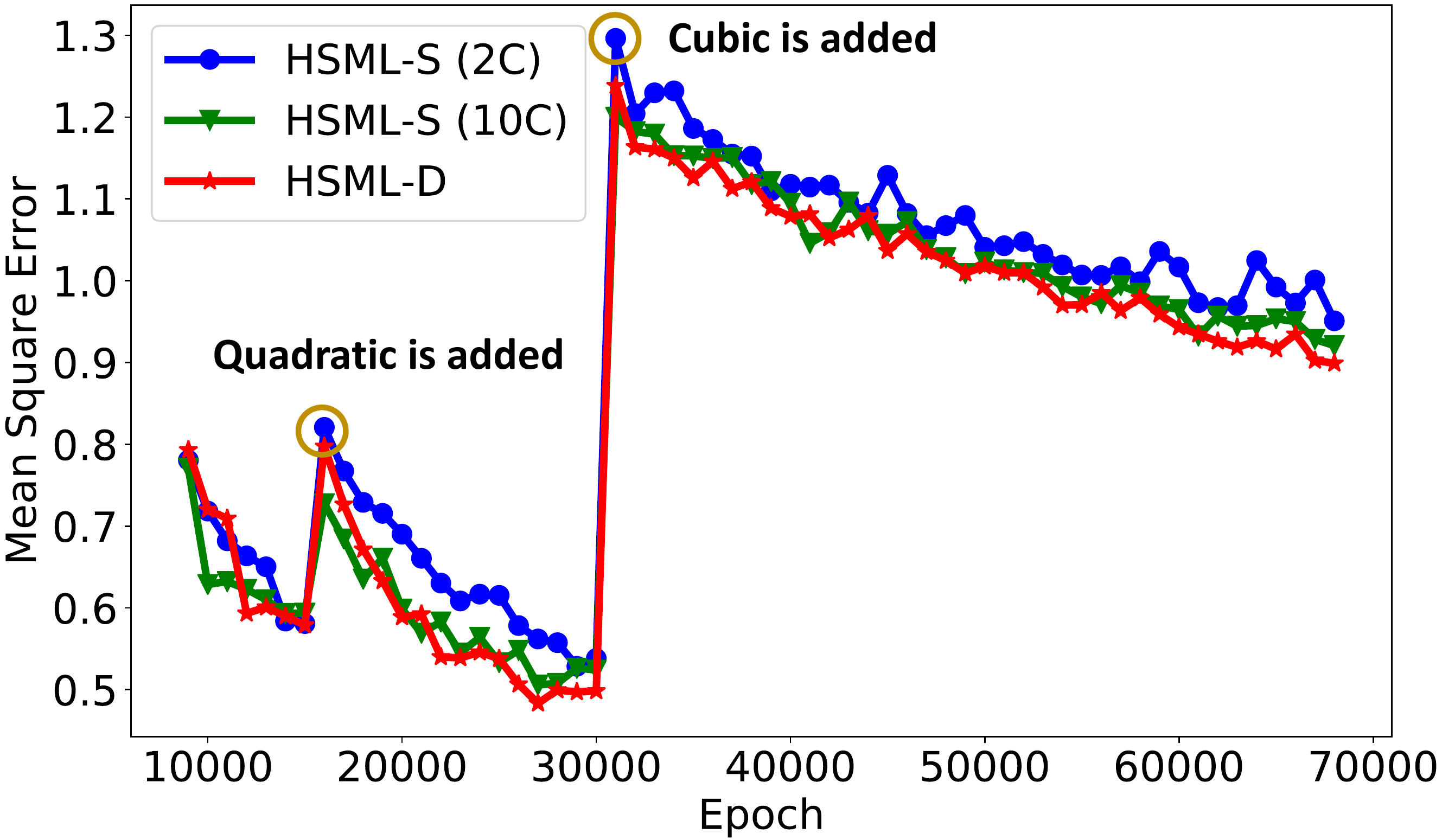}
	\end{subfigure}
	\begin{subtable}{0.95\linewidth}
	\scriptsize 
        \centering
        \begin{tabular}{lccc}
        \hline
           Model & HSML-S (2C) & HSML-S (10C) & HSML-D  \\
           MSE $\pm$ 95\% CI & $0.933\pm 0.074$ & $0.889\pm0.071$ & $\mathbf{0.869\pm0.072}$\\\hline
        \end{tabular}
    \end{subtable}  
	\caption{The performance comparison for the 5-shot toy regression problem in the continual adaptation scenario. The curve of MSE in meta-training process is shown in the top figure and the performance of meta-testing is reported in the bottom table.}
	\label{fig:sync_online}
\end{figure}

As shown in the heatmap, sinusoids and linear with positive slope activate cluster 1 and 3, respectively. Both quadratic 1 and 2 activate cluster 2, while quadratic 1 also activates cluster 1 and quadratic 2 also activates cluster 3. From the qualitative results, we can see the shape of quadratic 2 is similar to that of linear with positive slope, while quadratic 1 has more apparent curvature. Similar findings also verify in cubic cases. The shape of cubic 2 is very similar to sinusoids, thus cluster 1 is activated. Different from cubic 2, cubic 1 mainly activates cluster 4, whose shape is similar to linear with negative slope. The results indicate that the main cluster criteria of HSML is the shapes of tasks despite the underlying family functions. Furthermore, according to the qualitative results, HSML fits better than other baselines. 
\\
\textbf{Results of Continual Adaptation} 
To demonstrate the effectiveness of HSML under the continual learning scenario (HSML-D), we add more underlying functions during meta-training. 
First, we generate tasks from sinusoids and linear, and 
quadratic and cubic functions 
are
added after 15,000 and 30,000 training rounds, respectively. For comparison, one baseline 
is HSML 
with 2 
fixed clusters (HSML-S(2C)), and the other is HSML with 10 fixed clusters with much more representational capability (HSML-S(10C)). The meta-training loss curve and the meta-testing performance (MSE) are shown in Figure~\ref{fig:sync_online}. We can see that HSML-D outperforms as expected. Especially, HSML-D performs better than HSML-S(10C) which are prone to overfit and stuck at local optima at early stages.
\subsection{Few-shot Classification}
\textbf{Dataset and Experimental Settings}
In the few-shot classification problem, we construct a new benchmark which currently consists of four image classification datasets: Caltech-UCSD Birds-200-2011 (Bird)
~\cite{WahCUB_200_2011}, Describable Textures Dataset (Texture)~\cite{cimpoi14describing}, Fine-Grained Visual Classification of Aircraft (Aircraft)\mbox{~\cite{maji13fine-grained}}, and FGVCx-Fungi (Fungi)~\cite{Fungi} (See supplementary material B for detailed descriptions of this benchmark). Similar to the preprocessing of MiniImagenet~\cite{vinyals2016matching}, we divide each dataset to meta-training, meta-validation and meta-testing classes. Following the protocol in~\cite{finn2017model,ravi2016optimization,vinyals2016matching}, we adopt N-way classification with K-shot samples. Each task samples 
classes from one of the four datasets. Compared with previous benchmarks (e.g., MiniImagenet) that
the tasks are constructed within a single dataset, 
the new benchmark is more heterogeneous and closer to the real-world image classification. 
Like~\cite{finn2017model}, the base learner is a standard four-block convolutional architecture. The number of layers in hierarchical clustering structure is set as 3 with 4, 2, 1 clusters in each layer. Note that, in this section, for the tables without confidence interval, we provide the full results in supplementary material F. In addition, we provide the comparison to MiniImagenet benchmark in supplementary material D. Note that, the sampled tasks from MiniImagenet do not have obvious heterogeneity and uncertainty. Our approach achieves comparable results among gradient-based meta-learning methods.
\\
\textbf{Results of Classification Performance}
\begin{table*}[h]
\small
\caption{Comparison between HSML and other gradient-based meta-learning methods on the 5-way, 1-shot/5-shot image classification problem, averaged over 1000 tasks for each dataset. Accuracy $\pm$ $95\%$ confidence intervals are reported.}
\label{tab:metadataset_res}
\begin{center}
\small
\begin{tabular}{l|l|c|c|c|c|c}
\hline
& Model & Bird & Texture & Aircraft & Fungi & Average \\\hline
\multirow{6}{*}{\shortstack{5-way\\1-shot}} & MAML & $53.94\pm 1.45\%$ & $31.66\pm1.31\%$ & $51.37\pm 1.38\%$ & $42.12\pm 1.36\%$ & $44.77\%$ \\
& Meta-SGD & $55.58\pm1.43\%$ & $32.38\pm1.32\%$ & $52.99\pm1.36\%$ & $41.74\pm 1.34\%$ & $45.67\%$\\
& MT-Net & $58.72\pm 1.43\%$ & $32.80\pm1.35\%$ & $47.72\pm 1.46\%$ & $43.11\pm 1.42\%$ & $45.59\%$\\
& BMAML & $54.89\pm 1.48\%$ & $32.53\pm 1.33\%$ & $53.63\pm 1.37\%$ & $42.50\pm 1.33\%$ & $45.89\%$ \\
& MUMOMAML & $56.82\pm1.49\%$ & $33.81\pm1.36\%$ & $53.14\pm1.39\%$ & $42.22\pm1.40\%$ & $46.50\%$\\
\cmidrule{2-7}
& \textbf{HSML (ours)} & $\mathbf{60.98\pm1.50\%}$ & $\mathbf{35.01\pm 1.36\%}$ & $\mathbf{57.38\pm 1.40\%}$ & $\mathbf{44.02\pm 1.39\%}$ & $\mathbf{49.35\%}$\\\midrule\midrule
\multirow{6}{*}{\shortstack{5-way\\5-shot}} & MAML & $68.52\pm0.79\%$ & $44.56\pm0.68\%$ & $66.18\pm 0.71\%$ & $51.85\pm0.85\%$ & $57.78\%$ \\
& Meta-SGD & $67.87\pm0.74\%$ & $45.49\pm0.68\%$ & $66.84\pm0.70\%$ & $52.51\pm0.81\%$ & $58.18\%$\\
& MT-Net & $69.22\pm0.75\%$ & $46.57\pm0.70\%$ & $63.03\pm0.69\%$ & $53.49\pm0.83\%$ & $58.08\%$\\
& BMAML & $69.01\pm 0.74\%$ & $46.06\pm 0.69\%$ & $65.74\pm 0.67\%$ & $52.43\pm 0.84\%$ & $58.31\%$\\
& MUMOMAML & $70.49\pm0.76\%$ & $45.89\pm0.69\%$ & $67.31\pm0.68\%$ & $53.96\pm0.82\%$ & $59.41\%$\\
\cmidrule{2-7}
& \textbf{HSML (ours)} & $\mathbf{71.68\pm 0.73\%}$ & $\mathbf{48.08\pm 0.69\%}$ & $\mathbf{73.49\pm 0.68\%}$ & $\mathbf{56.32\pm 0.80\%}$ & $\mathbf{62.39\%}$\\\hline
\end{tabular}
\end{center}
\end{table*}
For each dataset, we report the averaged accuracy over 1000 tasks of 5-way 1-shot/5-shot classification in Table~\ref{tab:metadataset_res}. HSML consistently outperforms the other baselines on each dataset, which demonstrates the power of modeling hierarchical clustering structure. To verify the effectiveness of our proposed three components (i.e., task representation, hierarchical task clustering, knowledge adaptation), we also propose some variants of HSML. The detailed description of these variants and their corresponding results can be found in the supplementary material H, which further enhance the contribution of each component. In addition, we design another challenging leave-one-out experiment in this benchmark. We use one dataset for meta-testing and the rest three for meta-training. The results are reported in the supplementary material E and the HSML still achieves the best performance.
\\
\textbf{Task Clustering Analysis in Few-shot Classification}
Like the analysis of toy regression, we select four tasks in 5-way 1-shot classification and show their soft-assignment in Figure~\ref{fig:vis_metadata} (more results are shown in the supplementary material J). Darker color means higher probability. Furthermore, in Figure~\ref{fig:vis_metadata}, we show the learned hierarchical clustering of each task. In each layer, the top activated clusters are shown in darker color and then the activation paths are generated.
\begin{figure}[h]
	\centering
 	\includegraphics[width=0.98\linewidth]{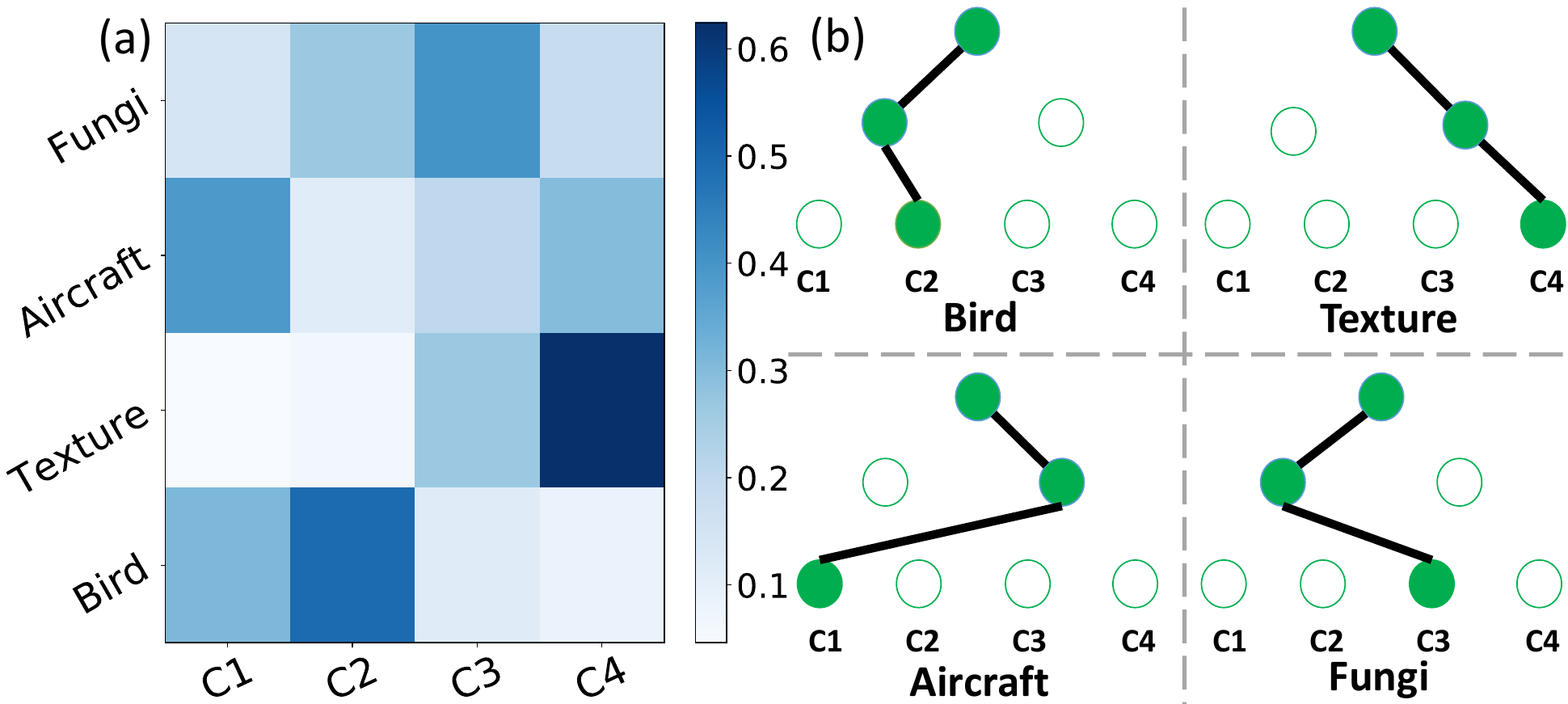}
	\caption{(a) The visualization of soft-assignment in Eqn.~\eqref{eq:cluster_assign} of four selected tasks.
	(b) Learned hierarchical structure of each task. In each layer, top activated cluster is shown in dark color.} 
	\label{fig:vis_metadata}
\end{figure}

\begin{figure}[h]
	\centering
	\begin{subfigure}[b]{0.235\textwidth}
		\centering
		\includegraphics[height=30mm]{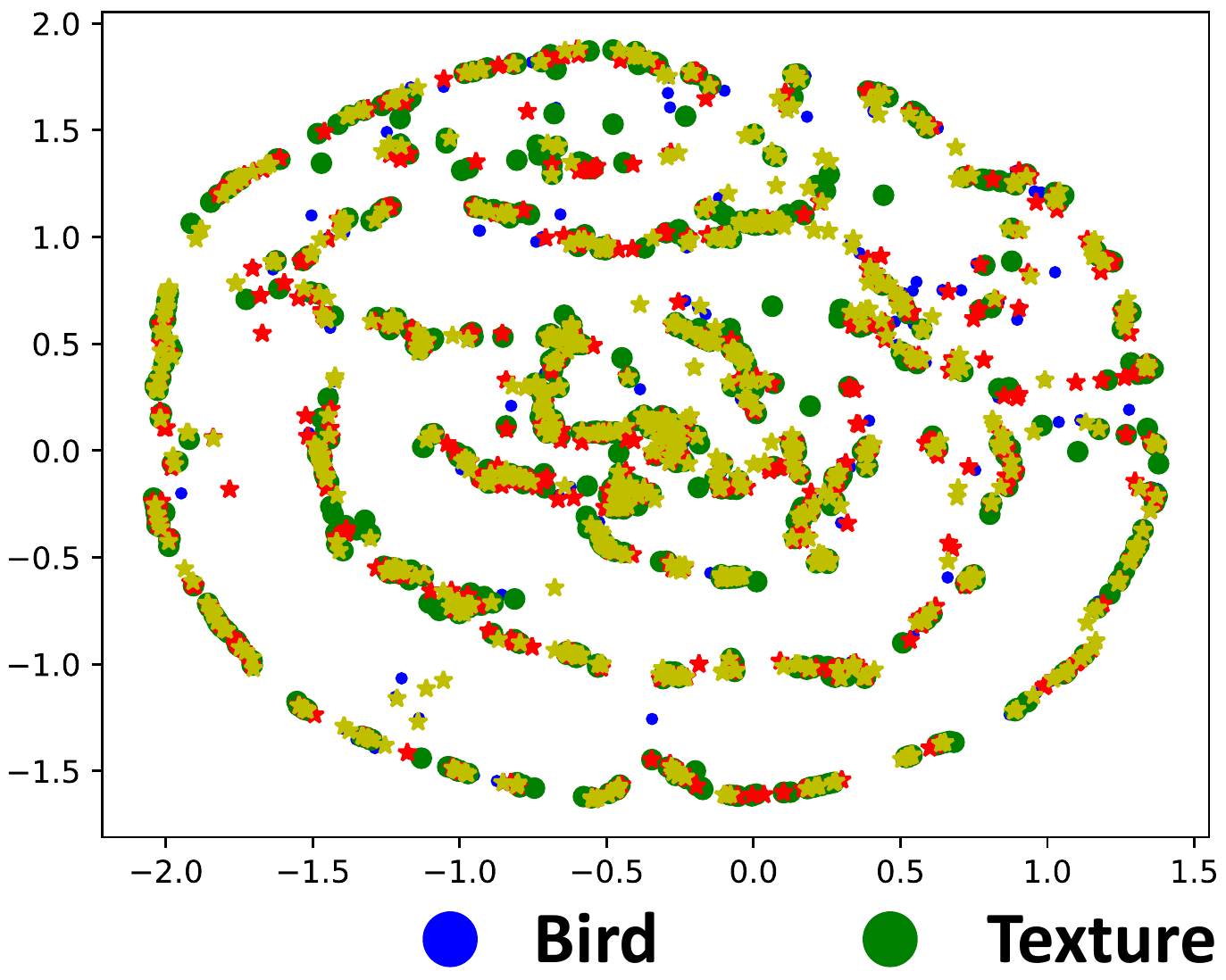}
		\caption{\label{fig:sinfunc}: MUMOMAML}
	\end{subfigure}
	\begin{subfigure}[b]{0.235\textwidth}
		\centering
		\includegraphics[height=30mm]{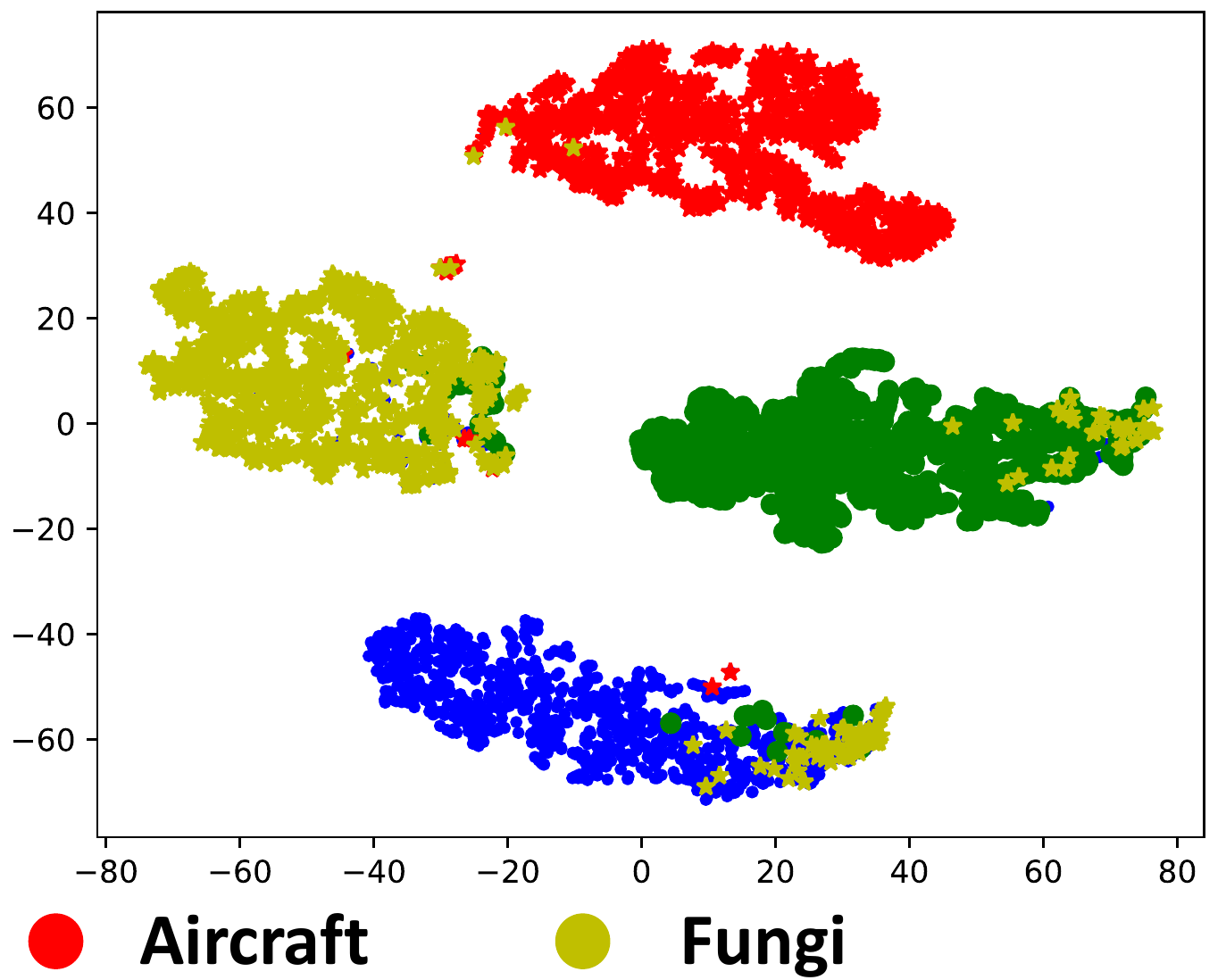}
		\caption{\label{fig:linefunc}: HSML (Ours)}
	\end{subfigure}
	\caption{t-SNE visualization of gated weight, i.e., $\theta_{0i}$, in Eqn.~\eqref{eq:lossall}}
	\label{fig:tsne_weight}
\end{figure}

\begin{figure}[h]
	\centering
	\begin{subfigure}[b]{0.85\linewidth}
	\centering
		\includegraphics[width=0.85\linewidth]{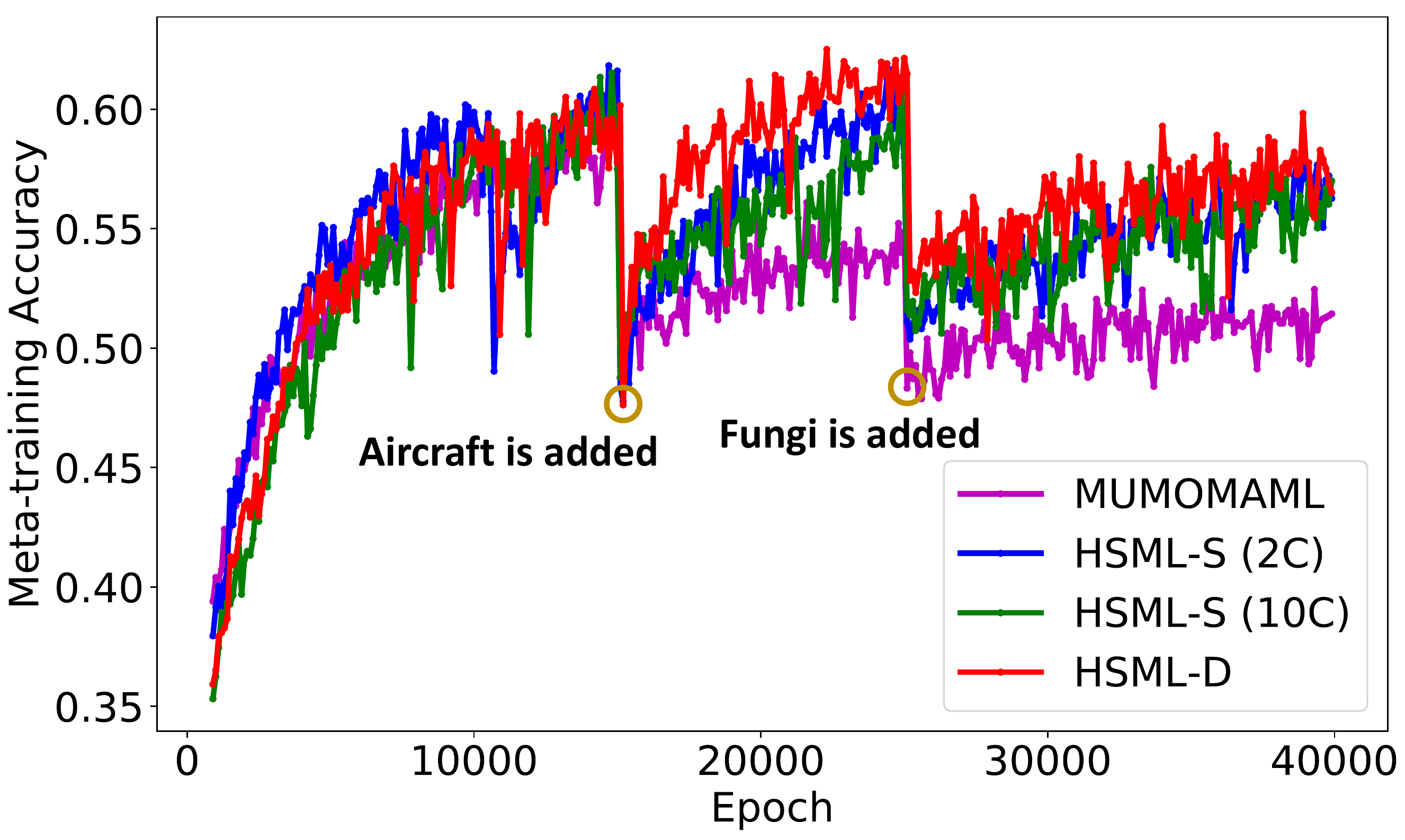}
	\end{subfigure}
	\begin{subtable}{0.95\linewidth}
	\small
        \centering
        \begin{tabular}{l|c|c|c|c}
        \hline
        Model & Bird & Texture & Aircraft & Fungi \\\hline
        MUMOMAML & 56.66\% & 33.68\% & 45.73\% & 40.38\% \\
        HSML-S (2C) & 60.77\% & 33.41\% & 51.28\% & 40.78\%\\
        HSML-S (10C) & 59.16\% & 34.48\% & 52.30\% & 40.56\%\\
        HSML-D & \textbf{61.16\%} & \textbf{34.53\%} & \textbf{54.50\%} & \textbf{41.66\%}\\\hline
        \end{tabular}
    \end{subtable}  
	\caption{The performance comparison for the 5-way 1-shot few-shot classification problem in the continual adaptation scenario. The top figure and bottom table show the meta-training accuracy curves and the meta-testing accuracy, respectively.} 
	\label{fig:metadataset_online}
\end{figure}

From Figure~\ref{fig:vis_metadata}, we can see different datasets mainly activate different clusters: bird$\rightarrow$cluster 2, texture$\rightarrow$cluster 4, aircraft$\rightarrow$cluster 1, fungi$\rightarrow$cluster 3. It is also interesting to find the clustering across different tasks via the second largest activated cluster which further promote knowledge transfer between tasks. The correlation may represent the similarity of shape (bird and aircraft), environment (fungi and bird), surface texture (texture and fungi). Note that, aircraft is correlated to texture because the classification of aircraft variant is mainly based on their shape and texture. The clustering can be further verified in the learned activated path. In the second layer, the left node, which may represent the environment, is activated by cluster 2 (activated by bird) and 3 (activated by fungi). The right node that reflects surface texture is activated by cluster 1 (activated by aircraft) and 4 (activated by texture). In Figure~\ref{fig:tsne_weight}, in addition, we randomly select 1000 tasks from each dataset, and show the t-SNE~\cite{maaten2008visualizing} visualization of the gated weight, i.e., $\theta_{0i}$, in Eqn.~\eqref{eq:lossall}. Compared with MUMOMAML, the results indicate that our clustering structure are able to identify the tasks in different clusters.
\\
\textbf{Results of Continual Adaptation}
In few-shot classification task, we conduct the experiments for continual adaptation in the 5-way 1-shot scenario. Initially, the tasks are generated from bird and texture datasets. Then, aircraft and fungi datasets are added after approximately meta-training round 15000 and 25000, respectively. We show the average meta-training accuracy curve and meta-testing accuracy in Figure~\ref{fig:metadataset_online}, where MUMOMAML, HSML-S(2C) and HSML-S(10C) are used as baselines. As shown in Figure~\ref{fig:metadataset_online}, HSML-D 
consistently achieves better performance.
\\
\textbf{Effect of Cluster Numbers}
We further analyze the effect of cluster numbers. The results are shown in Table~\ref{tab:cluster_sensitivity}. The cluster numbers from bottom layer to top layer are saved in a tuple. We can see that too few clusters may not enough to learn the task clustering characteristic (e.g., case (2,2,1)). In this dataset, increasing layers (e.g., case (8,4,4,1)) 
achieves similar performance compared with case (4,2,1). However, the former introduces more parameters.
\begin{table}[h]
\caption{Comparison of different cluster numbers. The numbers in first column represents the number of clusters from bottom layer to top layer. Accuracy for 5-way 1-shot classification are reported.} 
\label{tab:cluster_sensitivity}
\footnotesize
\begin{center}
\begin{tabular}{l|c|c|c|c}
\hline
Num. of Clu. & Bird & Texture & Aircraft & Fungi\\\hline
($2,2,1$) & $58.37\%$ & $33.18\%$ & $56.15\%$ & $42.90\%$\\
($4,2,1$) & $\mathbf{60.98\%}$ & $\mathbf{35.01\%}$ & $57.38\%$ & $44.02\%$\\
($6,3,1$) & $60.55\%$ & $34.02\%$ & $55.79\%$ & $43.43\%$\\
($8,4,2,1$) & $59.55\%$ & $34.74\%$ & $\mathbf{57.84\%}$ & $\mathbf{44.18\%}$ \\\hline
\end{tabular}
\end{center}
\end{table}
\section{Conclusion and Discussion}
In this paper, we introduce HSML to improve the meta-learning effectiveness, which simultaneously customizing task knowledge and preserving knowledge generalization via hierarchical clustering structure. Compared with several baselines, experiments demonstrated the effectiveness and interpretability of our algorithm in both toy regression and few-shot classification problems.

Although our method is widely applicable, there are some limitations and 
interesting 
future directions. (1) In this paper, we provide a simple version for continual learning, where tasks from new underlying groups are added continually. However, 
to construct a more reliable lifelong learning system, 
it is will be necessary to consider more complex evolution relations between tasks (e.g., relationship forgetting); (2) Another interesting direction is to combining active learning with task relation learning for automatically exploring evolutionary task relations.
\clearpage
\onecolumn
\appendix
\section*{Appendix}
\section{Detailed Theoretical Analysis}
\paragraph{Proof of Theorem 1}
Assuming a task $\mathcal{T}_i$ is sampled from $\mathcal{E}$, its training and testing samples are i.i.d. drawn from distribution $\mathcal{S}_i$, i.e., $\mathcal{D}_{\mathcal{T}_i}^{tr}\sim \mathcal{S}_i$ and $\mathcal{D}_{\mathcal{T}_i}^{te}\sim \mathcal{S}_i$. According to Theorem 3 in~\cite{kuzborskij2017data}, if $\mathcal{L}$ is convex, the base learner $f_{\theta_{\mathcal{T}_i}}$ SGD is $\epsilon(\mathcal{S}_i,\theta_0)$-on-average-stable with
\begin{equation}
\label{eq:app_convex}
    \epsilon(\mathcal{S}_i,\theta_0)=\mathcal{O}\bigg(\sqrt{c(R(\theta_0)-R^{*})}\frac{\sqrt[4]{T}}{n^{tr}}+c\sigma\frac{\sqrt{T}}{n^{tr}}\bigg),
\end{equation}
where $R^{*}=\inf_{\theta\in \mathcal{H}}R(\theta)$.

For a new task $\mathcal{T}_t$, we first prove that the initialization can be approximately represented as 
$\theta_{0t} = \sum_{k=1}^{K}\hat{\mathbf{B}}_k\theta_0$.
Wihtout loss of generality, here we consider a hierarchy $C-L-1$ in HSML. 

\begin{align}
    \theta_{0t} &= \theta_0 \circ \mathbf{o}_t \nonumber \\
     &= \text{diag}(\mathbf{o}_t)\theta_0 \nonumber \\
     & = \text{diag}(\text{FC}^{\sigma}_{\mathbf{W}_g}(\mathbf{g}_t\oplus\mathbf{h}_t))\theta_0 \nonumber \\
     & = \text{diag}(\text{FC}^{\sigma}_{\mathbf{W}_g}(\mathbf{g}_t\oplus\mathbf{h}_t))\theta_0 \nonumber \\
      & \approx \text{diag} \lbrace a_1[\mathbf{W}_g(\mathbf{g}_t\oplus\mathbf{h}_t)]+a_2\rbrace\theta_0 \nonumber \\
      & = \text{diag} [\mathbf{W}'_g(\mathbf{g}_t\oplus\mathbf{h}_t)+a_2]\theta_0 \nonumber \\
      & = \text{diag}\bigg\lbrace \mathbf{W}'_{gg}\mathbf{g}_t\oplus \mathbf{W}'_{gh}\sum_{l=1}^L p^l \tanh\bigg[\mathbf{W}\big(\sum_{c=1}^C p^{cl}\tanh(\mathbf{W}^l \mathbf{h}^c_t +b^l)\big) +b \bigg] + a_2\bigg\rbrace \nonumber \\
      & \approx \text{diag} \bigg\lbrace \mathbf{W}'_{gg}\mathbf{g}_t\oplus \mathbf{W}'_{gh} \sum_{l=1}^L p^l \bigg[\mathbf{W}\big(\sum_{c=1}^C p^{cl}(\mathbf{W}^l \mathbf{h}^c_t +b^l)\big) +b \bigg] + a_2\bigg\rbrace \nonumber \\
      & = \text{diag} \bigg\lbrace\sum_{l=1}^L\sum_{c=1}^C\bigg[\frac{1}{LC}\mathbf{W}'_{gg}\mathbf{g}_t\oplus p^l\mathbf{W}'_{gh}\bigg(p^{cl}\mathbf{W}\mathbf{W}^l\mathbf{h}^c_t + p^{cl}\mathbf{W}b^l + \frac{b}{C}\bigg) + \frac{a_2}{LC}\bigg]\bigg\rbrace\theta_0 \nonumber \\
       & = \sum_{k=1}^K \hat{\mathbf{B}}_k\theta_0,
\end{align}
where $K=CL$ and $\hat{\mathbf{B}}_{(l-1)*C+c}=\frac{1}{LC}\mathbf{W}'_{gg}\mathbf{g}_t\oplus p^l\mathbf{W}'_{gh}\bigg(p^{cl}\mathbf{W}\mathbf{W}^l\mathbf{h}^c_t + p^{cl}\mathbf{W}b^l + \frac{b}{C}\bigg) + \frac{a_2}{LC}$.
Note that the first equality holds by converting the Hadamard product into matrix multiplication, and the first and the second approximations come from first-order taylor series of sigmoid and hybolic functions. In addition, in the $C-L-1$ hierarchical structure, $\forall l$, $p^l=1$.

From Eqn.~\ref{eq:app_convex}, we can see that $\epsilon(\mathcal{S}_{t},\theta_0)$ depends $\sqrt{R(\theta_0)}$. Like~\cite{kuzborskij2017data}, when the optimization process for task $\mathcal{T}_t$ starts from the equivalent form that $\theta_{0t}=\sum_{k=1}^K \hat{\mathbf{B}}_k\theta_0$, we can bound $\epsilon(\mathcal{S}_{t},\theta_{0t})$ by using Hoeffding bound as:
\begin{equation}
\epsilon(\mathcal{S}_{t},\theta_{0t})\leq \mathcal{O}\bigg(\sqrt{\hat{R}_{\mathcal{D}^{tr}_{\mathcal{T}_t}}(\theta_{0t})+\sqrt{\frac{1}{n^{tr}}}}\bigg).
\end{equation}
Thus, we reach the conclusion.
\paragraph{Proof of Theorem 2}
In non-convex case, we assume $\mathcal{L}$ is $\eta$-smooth and has $\rho$-Lipschitz Hessian. According to the Corollary 1 and Proposition 1 in~\cite{kuzborskij2017data}, for task $\mathcal{T}_t$, we define:
\begin{equation}
    \gamma=\mathcal{O}\Bigg(\mathbb{E}_{(\mathbf{x}_{t,j},\mathbf{y}_{t,j})\sim \mathcal{D}_{\mathcal{T}_t}^{tr}}[\Vert\nabla^2\mathcal{L}(\theta_{0t},(\mathbf{x}_{t,j},\mathbf{y}_{t,j}))\Vert_2]+\sqrt{\hat{R}(\theta_{0t})}\Bigg),
\end{equation}
and
\begin{equation}
    \hat{\gamma}=\frac{1}{{n^{tr}}}\sum_{j=1}^{n^{tr}}\Vert\nabla^2\mathcal{L}(\theta_{0t}, (\mathbf{x}_{t,j},\mathbf{y}_{t,j}))\Vert_2+\sqrt{\hat{R}_{\mathcal{D}^{tr}_{\mathcal{T}_t}}(\theta_{0t})}.
\end{equation}

\begin{figure*}[!t]
    \centering
    \includegraphics[width=0.6\textwidth]{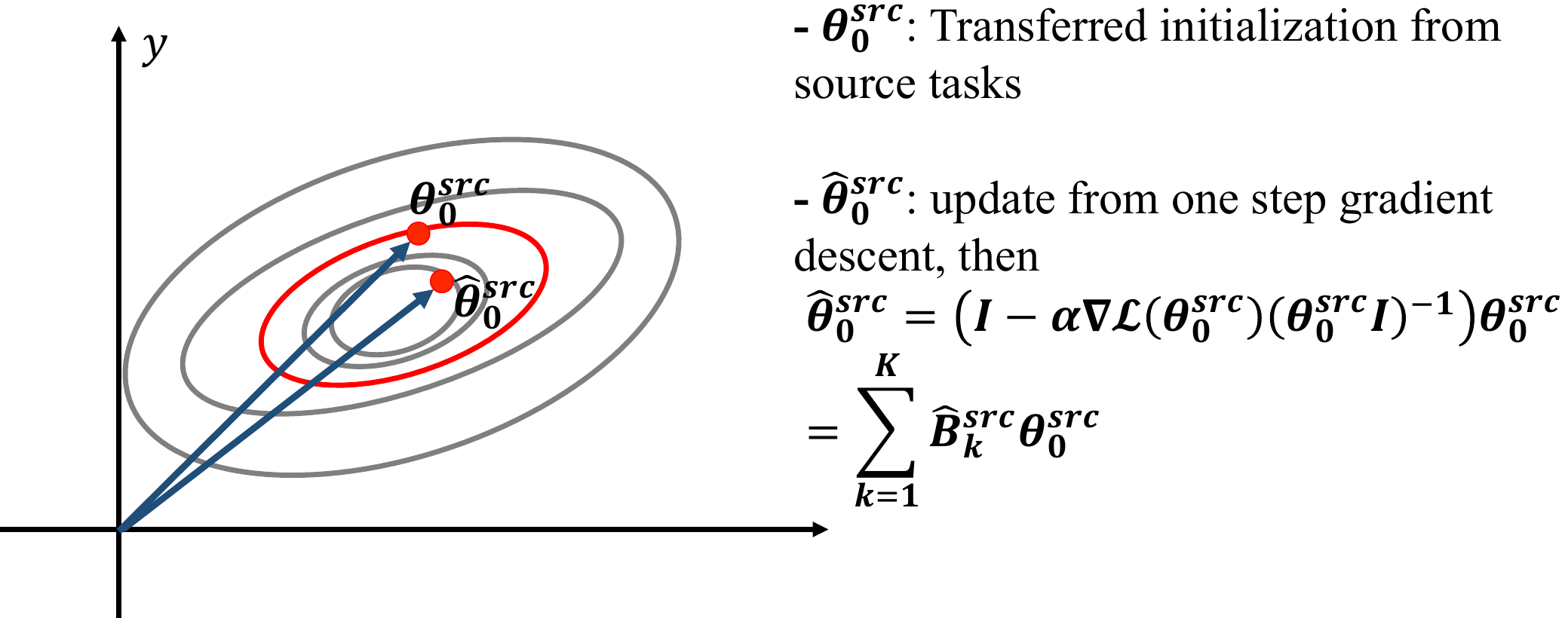}
    \caption{Illustration of Existance of $\sum_{k=1}^K \hat{\mathbf{B}}_k$.}
    \label{fig:gradient_app}
\end{figure*}

Then, we use Hoeffding inequality and get
\begin{equation}
    |\gamma-\hat{\gamma}|\leq \mathcal{O}(\frac{1}{\sqrt[4]{n^{tr}}}).
\end{equation}
Finally, let $\hat{\gamma}^{\pm}=\hat{\gamma}\pm 1/\sqrt[4]{n^{tr}}$, $\epsilon(\mathcal{S}_t,\theta_{0t})$ can be bounded as:
\begin{equation}
    \epsilon(\mathcal{S}_t,\theta_{0t})\leq \mathcal{O}\bigg(\Big(1+\frac{1}{c\hat{\gamma}^{-}}\Big)\hat{R}_{\mathcal{D}^{tr}_{\mathcal{T}_t}}(\theta_{0t})^{\frac{c\hat{\gamma}^{+}}{1+c\hat{\gamma}^{+}}}\frac{1}{{(n^{tr})}^{\frac{1}{1+c\hat{\gamma}^{+}}}}\bigg).
\end{equation}
Thus, we reach our conclusion.
\paragraph{Existance of $\sum_{k=1}^K \hat{\mathbf{B}}_k$} Here, we provides more details about the analysis of existence of $\sum_{k=1}^K \hat{\mathbf{B}}_k$, i.e., $\exists \{\hat{\mathbf{B}}_k\}_{k=1}^K, s.t.,\;\hat{R}_{\mathcal{D}^{tr}_{\mathcal{T}_t}}(\theta_{0t})\leq \hat{R}_{\mathcal{D}^{tr}_{\mathcal{T}_t}}(\theta_{0})$. Though the negative gradient descent, we can get 
\begin{equation}
\begin{aligned}
    \hat{\theta}_0&=\theta_0-\alpha\nabla\mathcal{L_{\theta_0}}\\
    &=(\mathbf{I}-\alpha\nabla\mathcal{L}(\theta_0)(\theta_0\mathbf{I})^{-1})\theta_0.
\end{aligned}
\end{equation}
Then, we can find a $\sum_{k=1}^K \hat{\mathbf{B}}_k=\mathbf{I}-\alpha\nabla\mathcal{L}(\theta_0)(\theta_0\mathbf{I})^{-1}$. It can also be verified in Figure~\ref{fig:gradient_app}. Assume $\theta_0$ is in the red contour, we can find a better parameter $\hat{\theta}_0$ inside the contour through its negative gradient direction. 
\section{Detailed Description of the New Few-shot Classification Benchmark}
\label{app:metadataset}
The new benchmark consists of four image classification datasets. All images are resized to $84\times84\times 3$. Here, we briefly introduce each of them as follows:
\begin{itemize}[leftmargin=*]
    \item \textbf{Caltech-UCSD Birds-200-2011 (CUB-200-2011)}~\cite{WahCUB_200_2011} is a bird image dataset which contains 11,788 photos of 200 bird species. In this paper, we randomly select 100 species with 60 photos in each species. We split the meta-training/meta-validation/meta-testing sets as 64/16/20 species. 
    \begin{itemize}
    \small
        \item \textbf{Meta-training:} Savannah Sparrow, Dark eyed Junco, Black footed Albatross, Henslow Sparrow, Cape Glossy Starling, Black throated Sparrow, Northern Waterthrush, Hooded Warbler, Baltimore Oriole, Scarlet Tanager, Cerulean Warbler, Downy Woodpecker, Black and white Warbler, Tropical Kingbird, Canada Warbler, Blue Jay, Elegant Tern, Groove billed Ani, Mallard, European Goldfinch, Red breasted Merganser, Geococcyx, Red winged Blackbird, Ringed Kingfisher, Prairie Warbler, Florida Jay, Hooded Oriole, American Redstart, Western Wood Pewee, Sayornis, Myrtle Warbler, Yellow Warbler, Tree Swallow, Rufous Hummingbird, Fish Crow, Bewick Wren, Seaside Sparrow, Vesper Sparrow, American Crow, Eared Grebe, Blue headed Vireo, White necked Raven, Frigatebird, Horned Lark, Tree Sparrow, Red bellied Woodpecker, Pacific Loon, Caspian Tern, Anna Hummingbird, Olive sided Flycatcher, Common Tern, Cedar Waxwing, Great Crested Flycatcher, Blue Grosbeak, White breasted Kingfisher, White eyed Vireo, Purple Finch, Cliff Swallow, Scissor tailed Flycatcher, Harris Sparrow, Western Grebe, Gadwall, American Goldfinch, Pine Warbler.
        \item \textbf{Meta-validation:} Mockingbird, Vermilion Flycatcher, Cape May Warbler, Prothonotary Warbler, White crowned Sparrow, Ovenbird, Pomarine Jaeger, Indigo Bunting, Blue winged Warbler, Chipping Sparrow, Horned Grebe, Fox Sparrow, Green Violetear, Nashville Warbler, Least Tern, Marsh Wren.
        \item \textbf{Meta-testing:} Rose breasted Grosbeak, Nighthawk, Long tailed Jaeger, Bronzed Cowbird, California Gull, Ivory Gull, Northern Fulmar, Brown Pelican, Ring billed Gull, Great Grey Shrike, White breasted Nuthatch, Mourning Warbler, Sage Thrasher, Horned Puffin, Pied Kingfisher, Shiny Cowbird, Scott Oriole, Red eyed Vireo, Song Sparrow, Winter Wren.
    \end{itemize}
    \item \textbf{Describable Textures Dataset (DTD)}~\cite{cimpoi14describing} is a texture image dataset which contains 5640 images from 47 classes. Each class contains 120 images. Meta-training/Meta-validation/Meta-testing contains 30/7/10 classes respectively. 
    \begin{itemize}
    \small
        \item \textbf{Meta-training:} pitted, woven, crosshatched, crystalline, sprinkled, lacelike, bubbly, marbled, dotted, bumpy, striped, zigzagged, lined, smeared, pleated, stratified, waffled, knitted, gauzy, porous, spiralled, grooved, banded, potholed, stained, veined, swirly, frilly, freckled, studded. 
        \item \textbf{Meta-validation:} wrinkled, grid, perforated, cobwebbed, honeycombed, cracked, blotchy.
        \item \textbf{Meta-testing:} fibrous, matted, scaly, chequered, flecked, paisley, braided, polka-dotted, interlaced, meshed.
    \end{itemize}
    \item \textbf{Fine-Grained Visual Classification of Aircraft (FGVC-Aircraft)}~\cite{maji13fine-grained} is a image dataset for fine grained visual categorization of aircraft. The dataset contains 102 different aircraft variants. In this paper, we randomly select 100 variants with 100 images in each variant. We split the meta-training/meta-validation/meta-testing to 64/16/20 variants respectively.
    \begin{itemize}
    \small
        \item \textbf{Meta-training:} MD-90, 737-600, A310, An-12, DR-400, Falcon-900, DC-3, Challenger-600, Fokker-70, Cessna-172, 747-400, ERJ-145, Dornier-328, A330-300, A319, Model-B200, E-170, A340-500, BAE-125, Metroliner, 747-300, C-130, DH-82, Hawk-T1, 727-200, 767-300, DC-10, Spitfire, E-195, BAE-146-300, F-16A-B, Beechcraft-1900, 747-200, Boeing-717, Falcon-2000, 777-300, Cessna-560, DHC-8-100, Cessna-525, 737-200, DC-8, Global-Express, DHC-1, CRJ-200, A340-300, DC-9-30, CRJ-900, A320, 737-300, Eurofighter-Typhoon, SR-20, E-190, Saab-340, C-47, Il-76, MD-87, 757-300, DHC-6, Tu-154, 777-200, 767-200, A318, 757-200, A300B4.
        \item \textbf{Meta-validation:} 737-900, A340-600, 737-800, 737-400, L-1011, A330-200, Gulfstream-V, 737-500, A340-200, ATR-72, MD-11, CRJ-700, EMB-120, Fokker-100, DC-6, 737-700.
        \item \textbf{Meta-testing:} 707-320, PA-28, Cessna-208, F-A-18, DHC-8-300, ERJ-135, Tornado, BAE-146-200, A321, ATR-42, Saab-2000, Tu-134, Fokker-50, A380, MD-80, Gulfstream-IV, Yak-42, 747-100, 767-400, Embraer-Legacy-600.
    \end{itemize}
    \item \textbf{FGVCx-Fungi (Fungi)}~\cite{Fungi} contains over 100,000 fungi images of nearly 1,500 wild mushroom species. We first filter the species with less than 150 images and then randomly select 100 species with 150 images in each species. We split the meta-training/meta-validation/meta-testing to 64/16/20 species respectively.
    \begin{itemize}
    \small
        \item \textbf{Meta-training}: Suillus granulatus, Phaeolus schweinitzii, Cystoderma amianthinum, Pycnoporellus fulgens, Psathyrella candolleana, Meripilus giganteus, Phellinus pomaceus, Laccaria laccata, Laccaria proxima, Amanita excelsa, Ganoderma pfeifferi, Clitopilus prunulus, Agaricus arvensis, Hericium coralloides, Plicatura crispa, Agrocybe praecox, Steccherinum ochraceum, Hypholoma fasciculare, Xerocomellus pruinatus, Xerocomellus chrysenteron, Crepidotus cesatii, Auricularia auricula-judae, Heterobasidion annosum, Entoloma clypeatum, Cortinarius torvus, Mycena tintinnabulum, Laetiporus sulphureus, Datronia mollis, Pholiota squarrosa, Cerioporus squamosus, Tricholoma terreum, Coprinellus micaceus, Cylindrobasidium laeve, Dacrymyces stillatus, Gloeophyllum sepiarium, Lycoperdon perlatum, Hygrophorus pustulatus, Clavulina coralloides, Xerocomus ferrugineus, Cortinarius alboviolaceus, Byssomerulius corium, Boletus edulis, Hymenopellis radicata, Basidioradulum radula, Cortinarius elatior, Schizophyllum commune, Cortinarius malicorius, Suillellus luridus, Ganoderma applanatum, Oligoporus guttulatus, Tubaria furfuracea, Cortinarius largus, Pleurotus ostreatus, Stereum hirsutum, Xylodon raduloides, Peniophora incarnata, Sutorius luridiformis, Flammulina velutipes var. velutipes, Phlebia radiata, Hygrocybe conica, Chlorophyllum olivieri, Armillaria ostoyae, Peniophora quercina, Mycena galericulata
        \item \textbf{Meta-validation:} Agaricus impudicus, Daedaleopsis confragosa, Fomitopsis pinicola, Cortinarius anserinus, Mucidula mucida, Trametes versicolor, Stropharia cyanea, Ramaria stricta, Radulomyces confluens, Gliophorus psittacinus, Psathyrella spadiceogrisea, Coprinopsis lagopus, Daedalea quercina, Amanita muscaria, Armillaria lutea, Vuilleminia comedens
        \item \textbf{Meta-testing:} Hygrocybe ceracea, Trametes hirsuta, Polyporus tuberaster, Lacrymaria lacrymabunda, Fistulina hepatica, Gymnopus dryophilus, Amanita rubescens, Fuscoporia ferrea, Craterellus undulatus, Tricholoma scalpturatum, Mycena pura, Russula depallens, Bjerkandera adusta, Trametes gibbosa, Tremella mesenterica, Cerioporus varius, Amanita fulva, Xylodon paradoxus, Cuphophyllus virgineus, Cortinarius flexipes
    \end{itemize}
\end{itemize}
\section{Hyperparameters \& Additional Experiment Settings}
We summarize the hyperparameters in this paper in Table~\ref{tab:parameters}. Like~\cite{finn2017model}, we compute the full Hessian-vector products for MAML. All cluster centers are randomly initialized. Note that, in few-shot classification problem, we use the change of averaged training accuracy to determine whether to increase clusters. Thus, $\mu<1$ in this problem. For toy regression task, the pre-aggregator embedding $\mathcal{F}(\cdot,\cdot)$ is a fully connected layer. Following~\cite{finn2017model}, the base learner has two hidden layers with 40 neurons in each. For few-shot image classification task, the pre-aggregator embedding $\mathcal{F}(\cdot,\cdot)$ is a block of two convolutional layers with two fully connected layers. The base learner is a standard base learner with 4 standard convolutional blocks. For continual scenario, we add one cluster every time. All the experiments are implemented using Tensorflow~\cite{abadi2016tensorflow}.
\begin{table*}[h]
\caption{Hyperparameter summary}
\label{tab:parameters}
\begin{center}
\begin{tabular}{l|c|c|c}
\hline
Hyperparameters & Toy Regreesion & miniImageNet & Multi-Datasets (New Benchmark) \\\hline
Input Scale (only for image data) & / & $84\times84\times 3$ & $84\times84\times 3$ \\
Meta-batch Size (task batch size) & 25 & 4 & 4\\
Inner loop learning rate ($\alpha$) & 0.001 & 0.001 & 0.001\\
Outer loop learning rate ($\beta$) & 0.001 & 0.01 & 0.01\\
Filters of CNN (only for image data) & / & 32 & 32\\
Meta-training adaptation steps & 5 & 5 & 5\\
Task representation size & 40 & 128 & 128\\
Reconstruction loss weight ($\gamma$) & 0.01 & 0.01 & 0.01\\
Image Embedding Size (before aggregator) & / & 64 & 64\\
Continual Training Threshold ($\tau$) & 1.25 & / & 0.85\\
\# epoch (Q) for computing loss & 1000 & / & 100 \\
\hline
\end{tabular}
\end{center}
\end{table*}
\label{app:para}

\section{Results of MiniImagenet}
\label{app:miniimagenet}
In this part, we present the additional comparison on MiniImagenet dataset. Similar to the analysis in~\cite{finn2018probabilistic}, the sampled tasks in this benchmark do not have obvious heterogeneity and uncertainty. Thus, the goal is to compare our approach with gradient-based meta-learning methods and other previous models. The expressive capacity of each model is controlled by using 4 standard convolutional layers and the results are shown in Table~\ref{tab:miniimagenet}. With the same expressive capacity, our model can achieve comparable performance with MAML-based models and other previous models in meta-learning field. 
\begin{table}[h]
\caption{Comparison between our approach and prior
few-shot learning techniques on the 5-way, 1-shot MiniImagenet benchmark. For MT-Net~\cite{lee2018gradient}, we remove the T-block since it introduces several $1\times 1$ convolutional layers which increases the expressive capacity of base learner~\cite{lin2013network}. For BMAML~\cite{yoon2018bayesian}, 24 classes are used for meta-testing in their original paper, while other methods use 20 classes. Since they have not released their code, we are not able to know the used classes. Thus, we implement it and report their performance on the standard classes (i.e., 20 classes for testing). Like~\cite{finn2018probabilistic}, we bold methods whose highest scores that overlap in their confidence intervals.}

\label{tab:miniimagenet}
\begin{center}
\begin{tabular}{l|c}
\hline
MiniImagenet & 5-way 1-shot Accuracy  \\\hline
Matching Nets~\cite{vinyals2016matching}    & $43.56\pm 0.34\%$  \\
meta-learner LSTM~\cite{ravi2016optimization}    & $43.44\pm 0.77\%$\\
Prototypical Network~\cite{snell2017prototypical}    & $46.61\pm 0.78\%$\\
SNAIL~\cite{mishra2018simple}    & $45.10\pm 0.00\%$\\
mAP-DLM~\cite{triantafillou2017few} & $49.82\pm 0.78\%$\\
Relation Net~\cite{yang2018learning}    & $\mathbf{50.44\pm 0.82\%}$\\
GNN~\cite{garcia2017few} & $\mathbf{50.33\pm 0.36\%}$ \\\midrule
MAML~\cite{finn2017model} & $48.70\pm1.84\%$\\
LLAMA~\cite{finn2017meta} & $49.40\pm 1.83\%$\\
BMAML~\cite{yoon2018bayesian} & $50.01\pm1.86\%$\\
MT-Net~\cite{lee2018gradient} & $49.75\pm1.83\%$\\
MUMOMAML~\cite{vuorio2018toward} & $49.86\pm 1.85\%$\\
Reptile~\cite{nichol2018reptile} & $49.97\pm0.32\%$\\
MetaSGD~\cite{li2017meta} & $\mathbf{50.47\pm 1.87\%}$\\
PLATIPUS~\cite{finn2018probabilistic} & $\mathbf{50.13\pm 1.86\%}$\\\midrule
\textbf{HSML (ours)}    & $\mathbf{50.38\pm 1.85\%}$  \\\hline
\end{tabular}
\end{center}
\end{table}

\section{Leave-one-out Experiments on Few-shot Image Classification}
\label{app:leave_one_out}
In this part, we design a more difficult experiment for few-shot image classification. For each dataset, we use three datasets for meta-training and the remaining dataset for meta-testing. For example, we use texture, bird and aircraft datasets for meta-training, and fungi dataset for meta-testing. Different from all the previous meta-learning settings which only use different classes for meta-testing, the leave-one-out experiment use a totally different dataset to test the generalization performance, which is more challenging. 

The results of 5-way 1-shot classification are shown in Table~\ref{tab:metadataset_res_leave_one_out}. We compare our methods with MAML and MUMOMAML (the best baseline in few-shot classification). We can see all results are significantly worse than the results without the leave-one-out technique, which shows the difficulty of this experiment. However, by capturing task clustering structure, our method can still achieves better performance than MAML and MUMOMAML.
\begin{table*}[!htbp]
\caption{Comparison of leave-one-out experiments on 5-way 1-shot classification. 4000 tasks are used to test the performance. For each dataset, the performance is reported when this dataset is used for meta-testing.}
\label{tab:metadataset_res_leave_one_out}
\begin{center}
\begin{tabular}{l|c|c|c|c|c}
\hline
Model & Bird & Texture & Aircraft & Fungi & Average \\\hline
MAML & $40.76\pm 0.68\%$ & $29.50\pm0.65\%$ & $29.54\pm0.63\%$ & $29.94\pm 0.64\%$ & $32.43\%$ \\
MUMOMAML & $41.58\pm 0.68\%$ & $30.24\pm 0.68\%$ & $30.69\pm0.66\%$ & $30.63\pm 0.66\%$ & $33.28\%$\\
HSML-RTG & $\mathbf{42.54 \pm 0.67\%}$ & $\mathbf{30.90\pm0.67\%}$ & $\mathbf{31.23\pm0.64\%}$ & $\mathbf{32.98\pm 0.68\%}$ & $\mathbf{34.41\%}$\\\hline
\end{tabular}
\end{center}
\end{table*}
\section{Additional Results of Few-shot Classification}
\label{app:additional_results}
Table~\ref{tab:metadataset_online_full} and Table~\ref{tab:app_cluster_sensitivity} contain the full results (accuracy with $95\%$ confident interval) of few-shot image classfiation. Table~\ref{tab:metadataset_online_full} shows the full results of the bottom table in Figure 7 (in paper). 
Table~\ref{tab:app_cluster_sensitivity} contains the full results of Table 3 (in paper).
\label{app:classification}
\begin{table*}[!htbp]
\caption{Comparison of online update results on few-shot image classification 5-way 1-shot scenario (Full Table).}
\label{tab:metadataset_online_full}
\begin{center}
\begin{tabular}{l|c|c|c|c|c}
\hline
Model & Bird & Texture & Aircraft & Fungi & Average \\\hline
MUMOMAML & $56.66\pm1.43\%$ & $33.68\pm1.37\%$ & $45.73\pm1.39\%$ & $40.38\pm1.40\%$ & $44.11\%$ \\
HSML-Static (2C) & $60.77\pm1.43\%$ & $33.41\pm 1.40\%$ & $51.28\pm 1.37\%$ & $40.78\pm 1.34\%$ & $46.56\%$\\
HSML-Static (10C) & $59.16\pm1.49\%$ & $34.48\pm 1.36\%$ & $52.30\pm 1.35\%$ & $40.56\pm 1.39\%$ & $46.63\%$\\
HSML-Dynamic & $\mathbf{61.16\pm 1.42\%}$ & $\mathbf{34.53\pm 1.35\%}$ & $\mathbf{54.50\pm 1.36\%}$ & $\mathbf{41.66\pm 1.41\%}$ & $\mathbf{47.96\%}$\\\hline
\end{tabular}
\end{center}
\end{table*}

\begin{table*}[!htbp]
\caption{Comparison of different cluster numbers (Full Table).}
\label{tab:app_cluster_sensitivity}
\begin{center}
\begin{tabular}{l|c|c|c|c|c}
\hline
Num. of Clus. & Bird & Texture & Aircraft & Fungi & Average\\\hline
($2,2,1$) & $58.37\pm 1.42\%$ & $33.18\pm 1.34\%$ & $56.15\pm 1.36\%$ & $42.90\pm 1.41\%$ & $47.65\%$ \\
($4,2,1$) & $\mathbf{60.98\pm1.50\%}$ & $\mathbf{35.01\pm 1.36\%}$ & $57.38\pm 1.40\%$ & $44.02\pm 1.39\%$ & $\mathbf{49.35\%}$ \\
($6,3,1$) & $60.55 \pm 1.45\%$ & $34.02\pm 1.34\%$ & $55.79\pm1.38\%$ & $43.43\pm1.39\%$ & $48.45\%$ \\
($8,4,4,1$) & $59.55\pm1.46\%$ & $34.74\pm1.37\%$ & $\mathbf{57.83\pm1.39\%}$ & $\mathbf{44.18\pm 1.38\%}$ & $49.08\%$ \\\hline

\end{tabular}
\end{center}
\end{table*}

\section{Effect of Different Aggregator}
\begin{table*}[!htbp]
\caption{Comparison of different aggregator on different shot, where HSML-RAA and HSML-MPAA represent HSML with recurrent autoencoder aggregator and mean pooling autoencoder aggregator, respectively.}
\label{tab:metadataset_aggregator_full}
\begin{center}
\begin{tabular}{l|l|c|c|c|c|c}
\hline
& Model & Bird & Texture & Aircraft & Fungi & Average \\\hline
\multirow{2}{*}{1-shot} & HSML-MPAA & $57.87\pm1.48\%$ & $32.07\pm1.36\%$ & $53.76\pm1.41\%$ & $40.88\pm1.37\%$ & $46.14\%$ \\
& HSML-RAA & $\mathbf{60.98\pm1.50\%}$ & $\mathbf{35.01\pm 1.36\%}$ & $\mathbf{57.38\pm 1.40\%}$ & $\mathbf{44.02\pm 1.39\%}$ & $\mathbf{49.35\%}$\\\midrule
\multirow{2}{*}{3-shot} & HSML-MPAA & $67.80\pm0.91\%$ & $44.33\pm0.82\%$ & $67.73\pm 0.83\%$ & $52.45\pm 0.94\%$ & $58.07\%$ \\
& HSML-RAA & $\mathbf{68.01\pm 0.88}\%$ & $\mathbf{45.07\pm 0.87\%}$ & $\mathbf{68.59\pm 0.82\%}$ & $\mathbf{53.51\pm 0.96\%}$ & $\mathbf{58.80\%}$\\\midrule
\multirow{2}{*}{5-shot} & HSML-MPAA & $\mathbf{71.80\pm0.70\%}$ & $48.02\pm0.68\%$ & $71.79\pm 0.74\%$ & $54.01\pm 0.82\%$ & $61.40\%$ \\
& HSML-RAA & $71.68\pm 0.73\%$ & $\mathbf{48.08\pm 0.69\%}$ & $\mathbf{73.49\pm 0.68\%}$ & $\mathbf{56.32\pm 0.80\%}$ & $\mathbf{62.39\%}$\\\midrule
\multirow{2}{*}{8-shot} & HSML-MPAA & $\mathbf{75.75\pm0.62\%}$ & $\mathbf{52.90\pm0.57\%}$ & $73.03\pm 0.55\%$ & $\mathbf{58.20\pm 0.73\%}$ & $64.97\%$ \\
& HSML-RAA & $75.52\pm 0.63\%$ & $51.52\pm 0.59\%$ & $\mathbf{75.33\pm 0.53\%}$ & $57.68\pm 0.71\%$ & $\mathbf{65.01\%}$\\\hline
\end{tabular}
\end{center}
\end{table*}
In our experiment, we found that the recurrent aggregator performs the best. To give more quantitative insight about the choice of aggregator, we compare these two aggregators with different shots in Table~\ref{tab:metadataset_aggregator_full}. We can see that recurrent aggregator significantly outperforms in 1-shot scenario. With the increase of the size of training samples, the performances of the two aggregators become more similar. Therefore, compared with recurrent aggregator, training a better mean pooling aggregator may require more data.

\section{Ablation Studies}
\label{app:ablation}
To investigate the contribution of different components of HSML (i.e., task representation, hierarchical task clustering, knowledge adaptation), we conduct the following ablation studies from four perspectives in Table~\ref{tab:ablationstudy}, where 5-way, 1-shot results on image classification are reported. The detailed ablations are provided in follows:
\begin{itemize}[leftmargin=*]
    \item (A1) We train 
    four MAMLs for 
    four clusters, i.e., bird, texture, aircraft and fungi, 
    by assigning a task to its groundtruth cluster. The results can be regarded as an upper-bound application of MAML with task clustering, provided with groundtruth clusters of all tasks which are unfortunately absent in real-world applications.
    HSML outperforms as the soft and hierarchical clustering not only accurately captures the task relationship but also encourages knowledge transfer across clusters.
    \item (A2) We investigate
    different variants of task representation learning in (A2a) and 
    (A2b). 
    In (A2a), we first use reconstruction loss to train  
    task embeddings. Next, we fix the parameters of the task representation learning component and backpropagate meta-gradients to only train the other two components. The results are inferior, showing that meta-learning gradients further optimize task embeddings. In (A2b), we replace our task embedding with 
    the last hidden state of the encoder. The results higher than 
    MUMOMAML 
    show the contribution of hierarchical clustering, while they worse than ours further justify the capability of our task embedding. 
    \item (A3) We analyze 
    the effect of hierarchical clustering in (A3). In (A3a), we remove the hierarchical task clustering  component. In (A3b), we consider the flat instead of hierarchical task clustering. 
    The results of (A3a) lower than (A3b) consolidate our motivation of knowledge generalization with a cluster.
    \item (A4) We also study three 
    variants of knowledge adaptation in (A4a)-(A4c). In (A4a), we revise Eqn. (8) by only using the  clustering representation. 
    The results still compete with state-of-the-art baselines, but empirically the combination with the task representation yields the best performance. In (A4b), we replace the parameter gate $\mathbf{o}_i$ with FiLM~\cite{perez2018film}, where the 
    comparable results verify the primary contribution of hierarchical clustering. In order to validate
    the effectiveness of parameter gate, in (A4c), we directly learn the initialization from 
    the task representation and the cluster representation instead of using
    the parameter gate to mask a shared initialization. The poor results show that the parameter gate masking a shared set of parameters $\theta_0$ may 
    1) prevent the curse of dimensionality and constrain the optimization space, given the high dimensionality of parameters; 
    2) serve as the warm-start for a new cluster of tasks in continual learning. 
\end{itemize}
\begin{table*}[!htbp]
\caption{Ablation Studies. Results of 5-way, 1-shot image classification are reported.}
\label{tab:ablationstudy}
\begin{center}
\begin{tabular}{p{150pt}|c|c|c|c}
\hline
Ablation & Bird & Texture & Aircraft & Fungi \\\hline
({\bf A1}): Train a MAML for each cluster, e.g., bird, by assigning a task to its groundtruth cluster.
&  $58.67\pm 1.49\%$ & $33.46\pm 1.34\%$ &  $55.81\pm 1.38\%$ & $43.50\pm 1.38\%$\\\midrule\midrule
({\bf A2a}): Use reconstruction loss to pretrain the task representation learning component, and then fix the paramters of it and backpropagate meta-gradients to only train the 
other two components.  &  $56.97\pm 1.44\%$ &  $29.12\pm 1.30\%$ &  $45.71\pm 1.38\%$ & $40.92\pm 1.39\%$\\\midrule
({\bf A2b}): Replace our task embedding with 
the last
hidden state of the encoder. & $58.25\pm 1.49\%$ &  $34.53\pm 1.36\%$ & $55.73\pm 1.37\%$ & $43.59\pm 1.39\%$\\\midrule\midrule
({\bf A3a}): Remove the hierarchical task clustering 
component.
& $58.22\pm 1.48\%$ & $33.30\pm 1.36\%$ & $55.35\pm 1.38\%$ &  $42.68\pm 1.40\%$\\\midrule
({\bf A3b}): Consider only flat rather than hierarchical task clustering. & $58.08\pm 1.45\%$ &  $34.26\pm 1.35\%$ & $56.11\pm 1.38\%$ &  $43.38\pm 1.39\%$\\\midrule\midrule
({\bf A4a}): 
Infer the parameter gate with the clustering representation only.
& $59.01\pm 1.50\%$ & $33.69\pm 1.35\%$ & $56.69\pm 1.39\%$ &  $42.88\pm 1.40\%$\\\midrule
({\bf A4b}): Replace the parameter gate with FiLM~\cite{perez2018film}. & $61.02\pm 1.47\%$ & $34.87\pm 1.37\%$ & $56.53\pm 1.40\%$ & $44.56\pm 1.38\%$ \\\midrule
({\bf A4c}): 
Learn the initialization directly from task and cluster representations rather than using the parameter gate.
& $53.95\pm 1.47\%$ & $32.35\pm 1.35\%$ & $52.15\pm 1.37\%$ & $42.31\pm 1.40\%$\\\hline

\end{tabular}
\end{center}
\end{table*}

\newpage
\section{Additional Task Clustering Results of Toy Regression Tasks}
\label{app:restoy}
In Figure~\ref{app:restoy}, we show the additional results of task clustering analysis of toy regression. In this figure, we further verify that tasks can be clustered by their shapes. Clusters 1 reflects the fluctuation mode curve (e.g., Sin a1-a4, Cubic a1-a4), while cluster 2 reflects an arc (e.g., Quad a2-a4). Cluster 3 mainly reflects a linear shape with positive slope (e.g. Line a1, Line a2, Quad a1, Quad a2, Cubic a1). Cluster 4 mainly reflects a linear shape with negative slope (e.g., Line a3, Line a4, Cubic a4). 
\begin{figure*}[!htbp]
    \centering
    \includegraphics[width=0.82\textwidth]{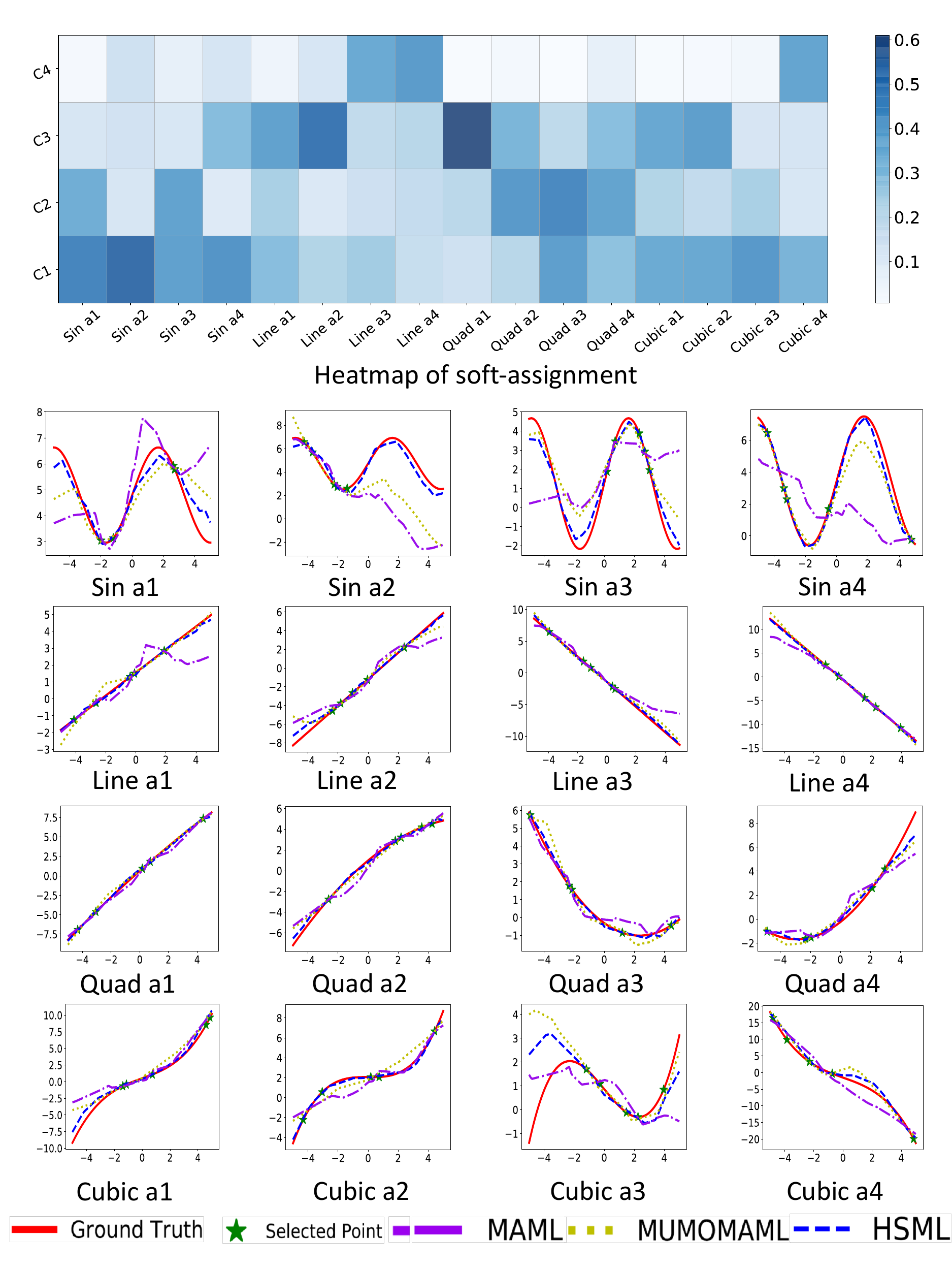}
    \caption{Additional results of task clustering analysis of toy regression problem.}
    \label{fig:toy_vis_app}
\end{figure*}

\section{Additional Task Clustering Analysis of Few-shot Classification}
\label{app:task_relation_classification}
\begin{figure*}[!htbp]
    \centering
    \includegraphics[width=0.80\textwidth]{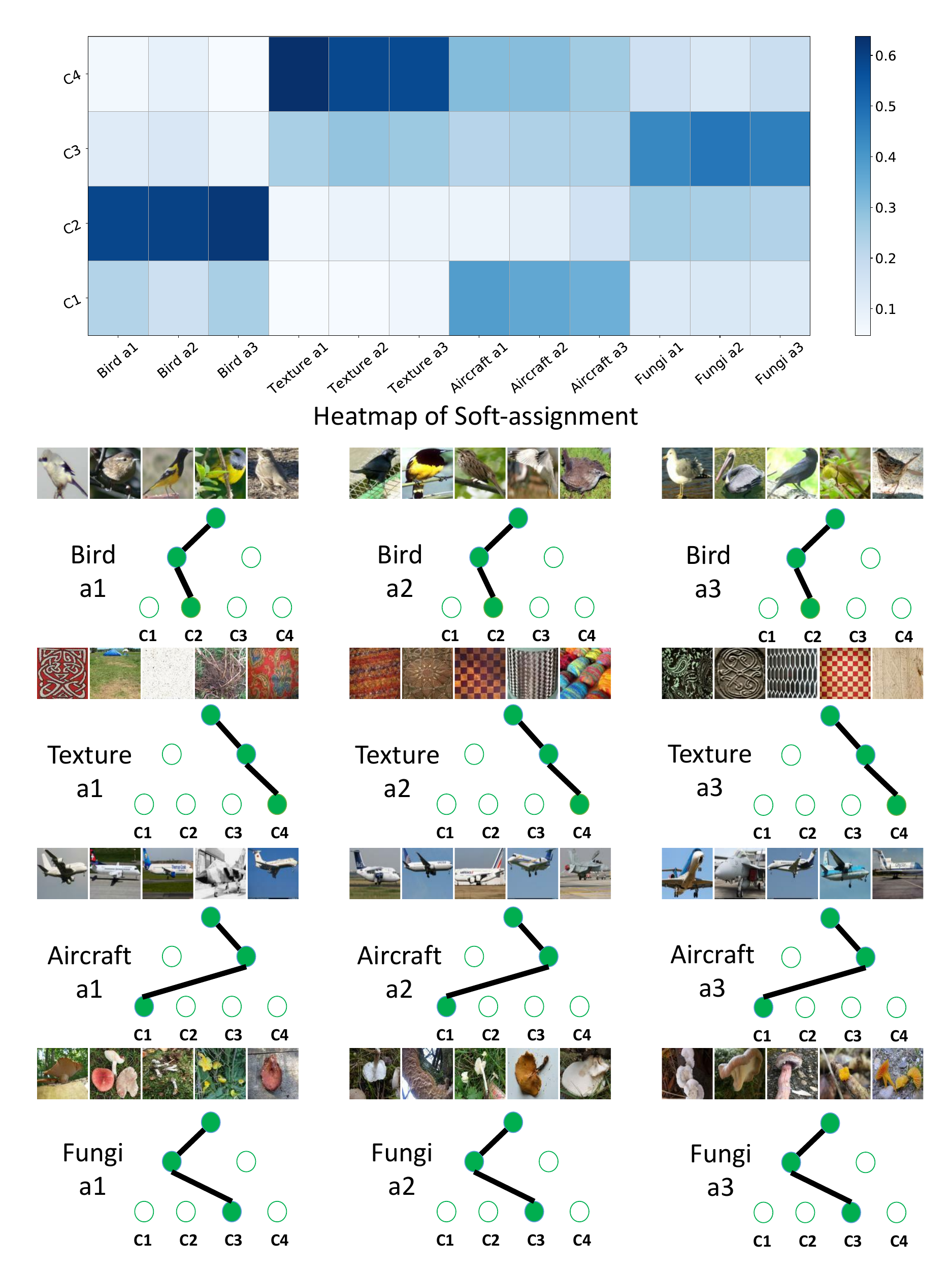}
    \caption{Additional results of task clustering analysis of few-shot image classification problem.}
    \label{fig:metadataset_vis_app}
\end{figure*}

In Figure~\ref{fig:metadataset_vis_app}, we show the additional results of task clustering analysis. The soft-assignment heatmap with their training images and activation paths of twelve tasks are illustrated. The conclusion is similar to that we draw previously in the paper. Tasks from different datasets mainly activate different clusters: bird$\rightarrow$cluster 2, texture$\rightarrow$cluster 4, aircraft$\rightarrow$cluster 1, fungi$\rightarrow$cluster 3. The left cluster and right cluster in the second layer may represent environment and surface texture, respectively. 

\clearpage
\twocolumn
\bibliography{ref}

\begin{thebibliography}{46}
\providecommand{\natexlab}[1]{#1}
\providecommand{\url}[1]{\texttt{#1}}
\expandafter\ifx\csname urlstyle\endcsname\relax
  \providecommand{\doi}[1]{doi: #1}\else
  \providecommand{\doi}{doi: \begingroup \urlstyle{rm}\Url}\fi

\bibitem[Fun(2018)]{Fungi}
2018 fgcvx fungi classification challenge, 2018.
\newblock URL \url{https://www.kaggle.com/c/fungi-challenge-fgvc-2018}.

\bibitem[Abadi et~al.(2016)Abadi, Barham, Chen, Chen, Davis, Dean, Devin,
  Ghemawat, Irving, Isard, et~al.]{abadi2016tensorflow}
Abadi, M., Barham, P., Chen, J., Chen, Z., Davis, A., Dean, J., Devin, M.,
  Ghemawat, S., Irving, G., Isard, M., et~al.
\newblock Tensorflow: a system for large-scale machine learning.
\newblock In \emph{OSDI}, volume~16, pp.\  265--283, 2016.

\bibitem[Andrychowicz et~al.(2016)Andrychowicz, Denil, Gomez, Hoffman, Pfau,
  Schaul, Shillingford, and De~Freitas]{andrychowicz2016learning}
Andrychowicz, M., Denil, M., Gomez, S., Hoffman, M.~W., Pfau, D., Schaul, T.,
  Shillingford, B., and De~Freitas, N.
\newblock Learning to learn by gradient descent by gradient descent.
\newblock In \emph{NIPS}, pp.\  3981--3989, 2016.

\bibitem[Baxter(1998)]{baxter1998theoretical}
Baxter, J.
\newblock Theoretical models of learning to learn.
\newblock In \emph{Learning to learn}, pp.\  71--94. Springer, 1998.

\bibitem[Braun et~al.(2010)Braun, Mehring, and Wolpert]{braun2010structure}
Braun, D.~A., Mehring, C., and Wolpert, D.~M.
\newblock Structure learning in action.
\newblock \emph{Behavioural brain research}, 206\penalty0 (2):\penalty0
  157--165, 2010.

\bibitem[Cimpoi et~al.(2014)Cimpoi, Maji, Kokkinos, Mohamed, , and
  Vedaldi]{cimpoi14describing}
Cimpoi, M., Maji, S., Kokkinos, I., Mohamed, S., , and Vedaldi, A.
\newblock Describing textures in the wild.
\newblock In \emph{CVPR}, 2014.

\bibitem[Conneau et~al.(2017)Conneau, Kiela, Schwenk, Barrault, and
  Bordes]{conneau2017supervised}
Conneau, A., Kiela, D., Schwenk, H., Barrault, L., and Bordes, A.
\newblock Supervised learning of universal sentence representations from
  natural language inference data.
\newblock In \emph{EMNLP}, pp.\  670--680, 2017.

\bibitem[Daniely et~al.(2015)Daniely, Gonen, and
  Shalev-Shwartz]{daniely2015strongly}
Daniely, A., Gonen, A., and Shalev-Shwartz, S.
\newblock Strongly adaptive online learning.
\newblock In \emph{ICML}, pp.\  1405--1411, 2015.

\bibitem[Finn \& Levine(2017)Finn and Levine]{finn2017meta}
Finn, C. and Levine, S.
\newblock Meta-learning and universality: Deep representations and gradient
  descent can approximate any learning algorithm.
\newblock \emph{arXiv preprint arXiv:1710.11622}, 2017.

\bibitem[Finn et~al.(2017)Finn, Abbeel, and Levine]{finn2017model}
Finn, C., Abbeel, P., and Levine, S.
\newblock Model-agnostic meta-learning for fast adaptation of deep networks.
\newblock In \emph{ICML}, pp.\  1126--1135, 2017.

\bibitem[Finn et~al.(2018)Finn, Xu, and Levine]{finn2018probabilistic}
Finn, C., Xu, K., and Levine, S.
\newblock Probabilistic model-agnostic meta-learning.
\newblock \emph{arXiv preprint arXiv:1806.02817}, 2018.

\bibitem[Flennerhag et~al.(2018)Flennerhag, Moreno, Lawrence, and
  Damianou]{flennerhag2018transferring}
Flennerhag, S., Moreno, P.~G., Lawrence, N.~D., and Damianou, A.
\newblock Transferring knowledge across learning processes.
\newblock \emph{arXiv preprint arXiv:1812.01054}, 2018.

\bibitem[Garcia \& Bruna(2017)Garcia and Bruna]{garcia2017few}
Garcia, V. and Bruna, J.
\newblock Few-shot learning with graph neural networks.
\newblock \emph{arXiv preprint arXiv:1711.04043}, 2017.

\bibitem[Gershman et~al.(2010)Gershman, Blei, and Niv]{gershman2010context}
Gershman, S.~J., Blei, D.~M., and Niv, Y.
\newblock Context, learning, and extinction.
\newblock \emph{Psychological review}, 117\penalty0 (1):\penalty0 197, 2010.

\bibitem[Gershman et~al.(2014)Gershman, Radulescu, Norman, and
  Niv]{gershman2014statistical}
Gershman, S.~J., Radulescu, A., Norman, K.~A., and Niv, Y.
\newblock Statistical computations underlying the dynamics of memory updating.
\newblock \emph{PLoS computational biology}, 10\penalty0 (11):\penalty0
  e1003939, 2014.

\bibitem[Grant et~al.(2018)Grant, Finn, Levine, Darrell, and
  Griffiths]{grant2018recasting}
Grant, E., Finn, C., Levine, S., Darrell, T., and Griffiths, T.
\newblock Recasting gradient-based meta-learning as hierarchical bayes.
\newblock \emph{arXiv preprint arXiv:1801.08930}, 2018.

\bibitem[Gu et~al.(2018)Gu, Wang, Chen, Cho, and Li]{gu2018meta}
Gu, J., Wang, Y., Chen, Y., Cho, K., and Li, V.~O.
\newblock Meta-learning for low-resource neural machine translation.
\newblock \emph{arXiv preprint arXiv:1808.08437}, 2018.

\bibitem[Hamilton et~al.(2017)Hamilton, Ying, and
  Leskovec]{hamilton2017inductive}
Hamilton, W., Ying, Z., and Leskovec, J.
\newblock Inductive representation learning on large graphs.
\newblock In \emph{NIPS}, pp.\  1024--1034, 2017.

\bibitem[Kim \& Xing(2010)Kim and Xing]{kim2010tree}
Kim, S. and Xing, E.~P.
\newblock Tree-guided group lasso for multi-task regression with structured
  sparsity.
\newblock In \emph{ICML}, pp.\  543--550, 2010.

\bibitem[Kumar \& Daum{\'e}~III(2012)Kumar and
  Daum{\'e}~III]{kumar2012learning}
Kumar, A. and Daum{\'e}~III, H.
\newblock Learning task grouping and overlap in multi-task learning.
\newblock In \emph{ICML}, pp.\  1723--1730, 2012.

\bibitem[Kuzborskij \& Lampert(2017)Kuzborskij and Lampert]{kuzborskij2017data}
Kuzborskij, I. and Lampert, C.~H.
\newblock Data-dependent stability of stochastic gradient descent.
\newblock \emph{arXiv preprint arXiv:1703.01678}, 2017.

\bibitem[Kuzborskij \& Orabona(2017)Kuzborskij and Orabona]{kuzborskij2017fast}
Kuzborskij, I. and Orabona, F.
\newblock Fast rates by transferring from auxiliary hypotheses.
\newblock \emph{Machine Learning}, 106\penalty0 (2):\penalty0 171--195, 2017.

\bibitem[Lee \& Choi(2018)Lee and Choi]{lee2018gradient}
Lee, Y. and Choi, S.
\newblock Gradient-based meta-learning with learned layerwise metric and
  subspace.
\newblock In \emph{ICML}, pp.\  2933--2942, 2018.

\bibitem[Li \& Malik(2016)Li and Malik]{li2016learning}
Li, K. and Malik, J.
\newblock Learning to optimize.
\newblock \emph{arXiv preprint arXiv:1606.01885}, 2016.

\bibitem[Li et~al.(2017)Li, Zhou, Chen, and Li]{li2017meta}
Li, Z., Zhou, F., Chen, F., and Li, H.
\newblock Meta-sgd: Learning to learn quickly for few shot learning.
\newblock \emph{arXiv preprint arXiv:1707.09835}, 2017.

\bibitem[Lin et~al.(2013)Lin, Chen, and Yan]{lin2013network}
Lin, M., Chen, Q., and Yan, S.
\newblock Network in network.
\newblock \emph{arXiv preprint arXiv:1312.4400}, 2013.

\bibitem[Maaten \& Hinton(2008)Maaten and Hinton]{maaten2008visualizing}
Maaten, L. v.~d. and Hinton, G.
\newblock Visualizing data using t-sne.
\newblock \emph{JMLR}, 9\penalty0 (Nov):\penalty0 2579--2605, 2008.

\bibitem[Maji et~al.(2013)Maji, Kannala, Rahtu, Blaschko, and
  Vedaldi]{maji13fine-grained}
Maji, S., Kannala, J., Rahtu, E., Blaschko, M., and Vedaldi, A.
\newblock Fine-grained visual classification of aircraft.
\newblock Technical report, 2013.

\bibitem[Mishra et~al.(2018)Mishra, Rohaninejad, Chen, and
  Abbeel]{mishra2018simple}
Mishra, N., Rohaninejad, M., Chen, X., and Abbeel, P.
\newblock A simple neural attentive meta-learner.
\newblock \emph{ICLR}, 2018.

\bibitem[Munkhdalai \& Yu(2017)Munkhdalai and Yu]{munkhdalai2017meta}
Munkhdalai, T. and Yu, H.
\newblock Meta networks.
\newblock In \emph{ICML}, pp.\  2554--2563, 2017.

\bibitem[Munkhdalai et~al.(2018)Munkhdalai, Yuan, Mehri, and
  Trischler]{munkhdalai2018rapid}
Munkhdalai, T., Yuan, X., Mehri, S., and Trischler, A.
\newblock Rapid adaptation with conditionally shifted neurons.
\newblock In \emph{ICML}, pp.\  3661--3670, 2018.

\bibitem[Nguyen \& Hein(2018)Nguyen and Hein]{nguyen2018optimization}
Nguyen, Q. and Hein, M.
\newblock Optimization landscape and expressivity of deep cnns.
\newblock In \emph{ICML}, 2018.

\bibitem[Nichol \& Schulman(2018)Nichol and Schulman]{nichol2018reptile}
Nichol, A. and Schulman, J.
\newblock Reptile: a scalable metalearning algorithm.
\newblock \emph{arXiv preprint arXiv:1803.02999}, 2018.

\bibitem[Perez et~al.(2018)Perez, Strub, de~Vries, Dumoulin, and
  Courville]{perez2018film}
Perez, E., Strub, F., de~Vries, H., Dumoulin, V., and Courville, A.~C.
\newblock Film: Visual reasoning with a general conditioning layer.
\newblock In \emph{AAAI}, 2018.

\bibitem[Ravi \& Larochelle(2016)Ravi and Larochelle]{ravi2016optimization}
Ravi, S. and Larochelle, H.
\newblock Optimization as a model for few-shot learning.
\newblock \emph{ICLR}, 2016.

\bibitem[Santoro et~al.(2016)Santoro, Bartunov, Botvinick, Wierstra, and
  Lillicrap]{santoro2016meta}
Santoro, A., Bartunov, S., Botvinick, M., Wierstra, D., and Lillicrap, T.
\newblock Meta-learning with memory-augmented neural networks.
\newblock In \emph{ICML}, pp.\  1842--1850, 2016.

\bibitem[Snell et~al.(2017)Snell, Swersky, and Zemel]{snell2017prototypical}
Snell, J., Swersky, K., and Zemel, R.
\newblock Prototypical networks for few-shot learning.
\newblock In \emph{NIPS}, pp.\  4077--4087, 2017.

\bibitem[Triantafillou et~al.(2017)Triantafillou, Zemel, and
  Urtasun]{triantafillou2017few}
Triantafillou, E., Zemel, R., and Urtasun, R.
\newblock Few-shot learning through an information retrieval lens.
\newblock In \emph{NIPS}, pp.\  2255--2265, 2017.

\bibitem[Vinyals et~al.(2016)Vinyals, Blundell, Lillicrap, Wierstra,
  et~al.]{vinyals2016matching}
Vinyals, O., Blundell, C., Lillicrap, T., Wierstra, D., et~al.
\newblock Matching networks for one shot learning.
\newblock In \emph{NIPS}, pp.\  3630--3638, 2016.

\bibitem[Vuorio et~al.(2018)Vuorio, Sun, Hu, and Lim]{vuorio2018toward}
Vuorio, R., Sun, S.-H., Hu, H., and Lim, J.~J.
\newblock Toward multimodal model-agnostic meta-learning.
\newblock \emph{arXiv preprint arXiv:1812.07172}, 2018.

\bibitem[Wah et~al.(2011)Wah, Branson, Welinder, Perona, and
  Belongie]{WahCUB_200_2011}
Wah, C., Branson, S., Welinder, P., Perona, P., and Belongie, S.
\newblock {The Caltech-UCSD Birds-200-2011 Dataset}.
\newblock Technical Report CNS-TR-2011-001, California Institute of Technology,
  2011.

\bibitem[Xu et~al.(2015)Xu, Ba, Kiros, Cho, Courville, Salakhudinov, Zemel, and
  Bengio]{xu2015show}
Xu, K., Ba, J., Kiros, R., Cho, K., Courville, A., Salakhudinov, R., Zemel, R.,
  and Bengio, Y.
\newblock Show, attend and tell: Neural image caption generation with visual
  attention.
\newblock In \emph{ICML}, pp.\  2048--2057, 2015.

\bibitem[Yang et~al.(2018)Yang, Zhang, Xiang, Torr, and
  Hospedales]{yang2018learning}
Yang, F. S.~Y., Zhang, L., Xiang, T., Torr, P.~H., and Hospedales, T.~M.
\newblock Learning to compare: Relation network for few-shot learning.
\newblock In \emph{CVPR}, 2018.

\bibitem[Ying et~al.(2018)Ying, You, Morris, Ren, Hamilton, and
  Leskovec]{ying2018hierarchical}
Ying, Z., You, J., Morris, C., Ren, X., Hamilton, W., and Leskovec, J.
\newblock Hierarchical graph representation learning with differentiable
  pooling.
\newblock In \emph{NIPS}, pp.\  4805--4815, 2018.

\bibitem[Yoon et~al.(2018{\natexlab{a}})Yoon, Kim, Dia, Kim, Bengio, and
  Ahn]{yoon2018bayesian}
Yoon, J., Kim, T., Dia, O., Kim, S., Bengio, Y., and Ahn, S.
\newblock Bayesian model-agnostic meta-learning.
\newblock In \emph{NIPS}, pp.\  7343--7353, 2018{\natexlab{a}}.

\bibitem[Yoon et~al.(2018{\natexlab{b}})Yoon, Yang, Lee, and
  Hwang]{yoon2017lifelong}
Yoon, J., Yang, E., Lee, J., and Hwang, S.~J.
\newblock Lifelong learning with dynamically expandable networks.
\newblock In \emph{ICLR}, 2018{\natexlab{b}}.

\end{thebibliography}
\bibliographystyle{icml2019}
\end{document}